\documentclass[journal]{IEEEtai}
\let\labelindent\relax
\usepackage[colorlinks,urlcolor=blue,linkcolor=blue,citecolor=blue]{hyperref}
\usepackage{color,array}
\usepackage{graphicx}
\usepackage{epsfig}
\usepackage{amsmath}
\usepackage{amssymb}

\usepackage{multirow}
\usepackage[table,xcdraw]{xcolor}
\usepackage{multirow}
\usepackage[normalem]{ulem}
\useunder{\uline}{\ul}{}
\usepackage{color}
\usepackage{graphicx}

\usepackage{booktabs}
\usepackage{stfloats}
\usepackage{graphicx}

\usepackage{booktabs}
\usepackage{colortbl}

\usepackage{multirow} 
\usepackage{overpic}
\usepackage{hhline}

\usepackage{arydshln} 
\usepackage{pifont}
\usepackage{enumitem}
\usepackage{colortbl}
\usepackage{makecell}
\usepackage{verbatim}
\usepackage{bm}
\usepackage{caption,subcaption}
\usepackage{bbding}
\usepackage{algorithm}
\usepackage{algorithmic}
\usepackage{multirow}

\begin{document}

\title{Brightness Perceiving for Recursive Low-Light Image Enhancement} 

\author{Haodian Wang, Long Peng, Yuejin Sun, Zengyu Wan, Yang Wang$^{\ast}$, Yang Cao$^{\ast}$, \IEEEmembership{Member, IEEE}
\thanks{This work was supported by the Natural Science Foundation of China No. 62206262, and National Key R\&D Program of China under Grant 2020AAA0105700. Haodian Wang and Long Peng are the co-first authors. $^{\ast}$Corresponding author: Yang Wang and Yang Cao.}
\thanks{Haodian Wang, Long Peng, Yuejin Sun, Zengyu Wan, Yang Wang, Yang Cao are with the University of Science and Technology of China, Anhui, China (e-mail: $\{$wanghaodian, longp2001, wanzengy, yjsun97$\}$@mail.ustc.edu.cn, $\{$ywang120, forrest$\}$@ustc.edu.cn).} \thanks{The related code will be released at 
\href{https://github.com/7407810/Brightness-Perceiving-for-Recursive-LLIE}{https://github.com/7407810/Brightness-Perceiving-for-Recursive-LLIE}.
}}

\markboth{Journal of IEEE Transactions on Artificial Intelligence}
{Wang \MakeLowercase{\textit{et al.}}: Brightness Perceiving for Recursive Low-Light Image Enhancement}

\maketitle

\begin{abstract}
Due to the wide dynamic range in real low-light scenes, there will be large differences in the degree of contrast degradation and detail blurring of captured images, making it difficult for existing end-to-end methods to enhance low-light images to normal exposure. To address the above issue, we decompose low-light image enhancement into a recursive enhancement task and propose a brightness-perceiving-based recursive enhancement framework for high dynamic range low-light image enhancement. Specifically, our recursive enhancement framework consists of two parallel sub-networks: Adaptive Contrast and Texture enhancement network (ACT-Net) and Brightness Perception network (BP-Net). The ACT-Net is proposed to adaptively enhance image contrast and details under the guidance of the brightness adjustment branch and gradient adjustment branch, which are proposed to perceive the degradation degree of contrast and details in low-light images. To adaptively enhance images captured under different brightness levels, BP-Net is proposed to control the recursive enhancement times of ACT-Net by exploring the image brightness distribution properties. Finally, in order to coordinate ACT-Net and BP-Net, we design a novel unsupervised training strategy to facilitate the training procedure.  
To further validate the effectiveness of the proposed method, we construct a new dataset with a broader brightness distribution by mixing three low-light datasets. Compared with eleven existing representative methods, the proposed method achieves new SOTA performance on six reference and no reference metrics. Specifically, the proposed method improves the PSNR by 0.9 dB compared to the existing SOTA method.
\end{abstract}

\begin{IEEEImpStatement}

The degrees of detail loss and contrast degradation vary in high-dynamic low-light scenarios, making it challenging to reconstruct high-quality images. To address this issue, we propose a recursive enhancement framework that gradually improves contrast and detail texture while utilizing a brightness perception branch to control the extent of enhancement. Extensive experiments demonstrate that our proposed method outperforms existing methods and achieves remarkable generalization performance in real-world scenarios. Consequently, our method possesses prospective applications on intelligent displays and cameras to improve model robustness in complex and variable lighting conditions.

\end{IEEEImpStatement}


\section{Introduction}

\begin{figure}
    \centering
    \includegraphics[width=1.0\linewidth]
    {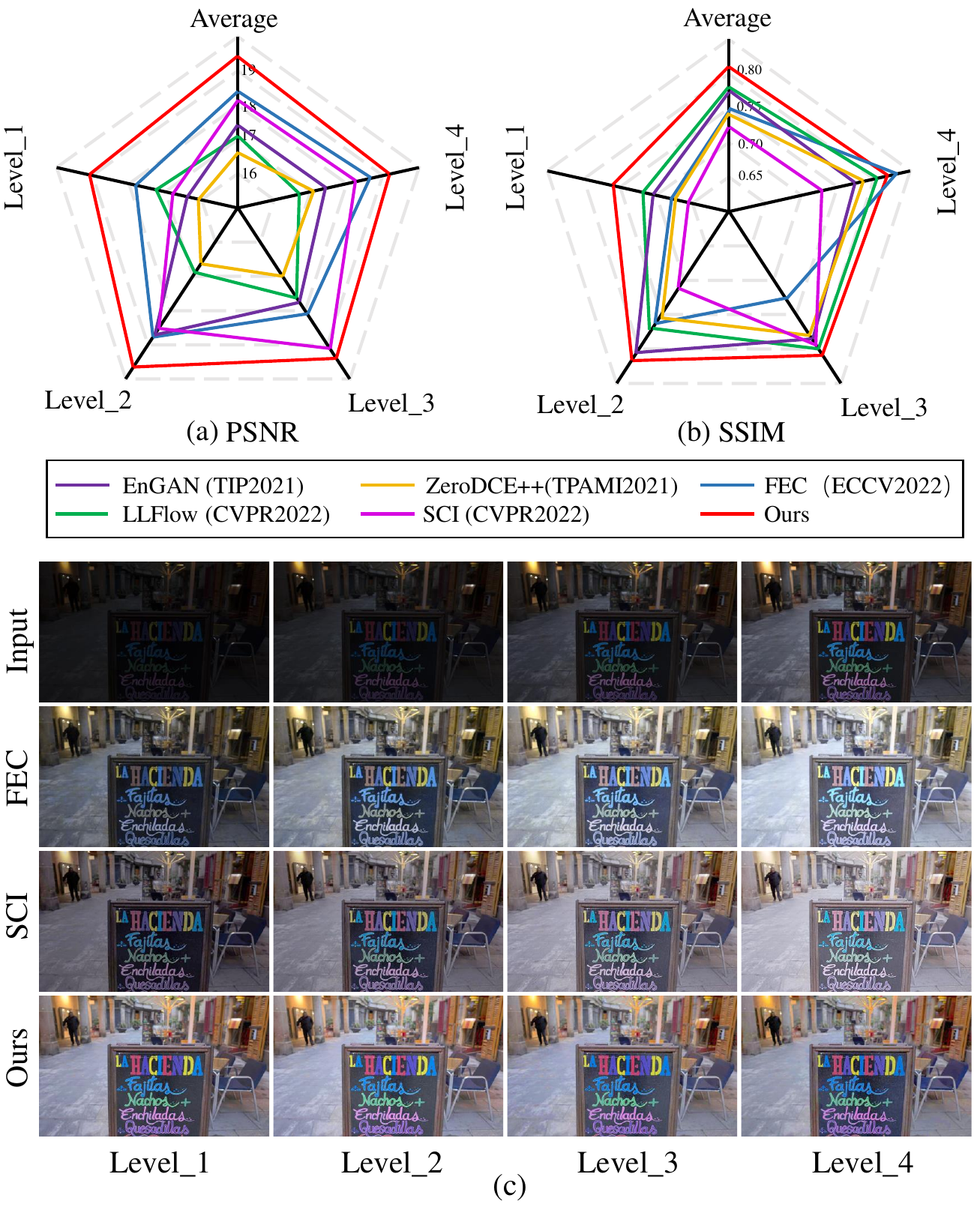}
    \caption{(a, b) Comparison with existing methods on PSNR and SSIM under different brightness levels. (c) Visualization comparison with existing 
representative multi-exposure enhancement method FEC \cite{huang2022deep} and low light enhancement method SCI \cite{ma2022toward} in 4 different brightness of the same background. }
    \label{fig:Fig1}
    
\end{figure}

\begin{figure*}
    \centering
    \includegraphics[width=1.0\linewidth]
    {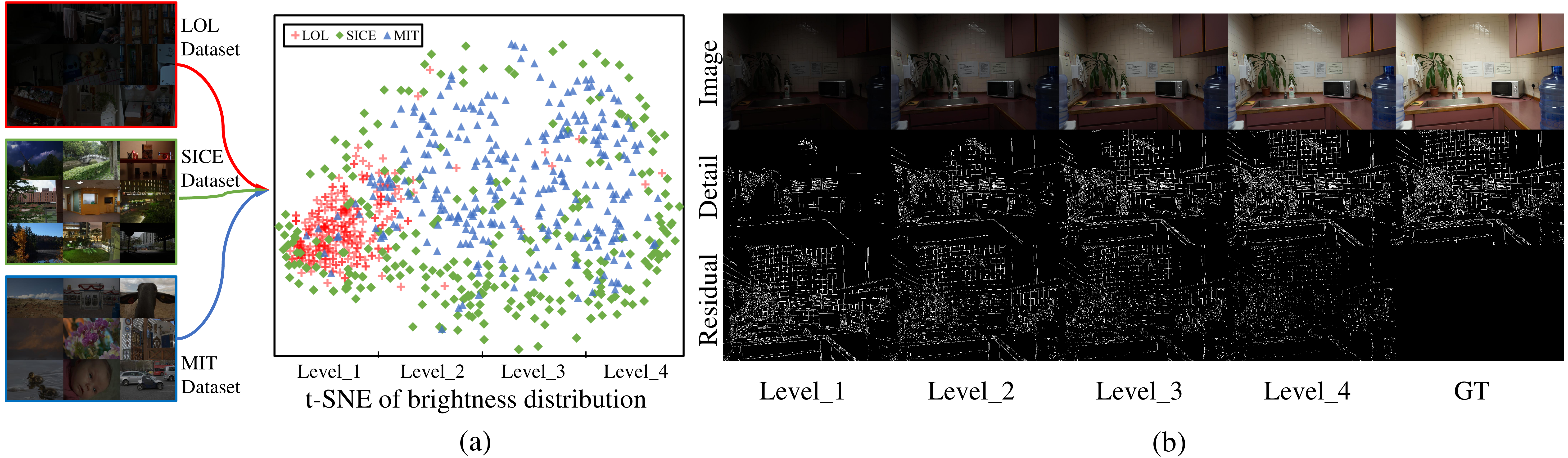}
    \setlength{\abovecaptionskip}{-0.15cm}
    \setlength{\belowcaptionskip}{-0.2cm}
    \caption{(a) The t-SNE \cite{van2008visualizing} of brightness distribution of existing three prevailing low-light datasets. (b) The visualization of details degradation in different low-light scenes from Level\_1 (darkest) to Level\_4 (brightest). We can observe that the brightness distribution of low-light images varies significantly under different brightness levels; the lower the brightness, the greater the structural damage.}
    \label{fig:Fig2}
\end{figure*}

\IEEEPARstart{T}{he} images captured under low illumination will suffer from statistical and structural properties distortion, resulting in low contrast, blur, and noise problems. This can significantly degrade the human visual effects and performance of high-level vision algorithms \cite{wang2020experiment,lu2022progressive,add_reference_1,add_reference_2,add_reference_3,add_reference_4}. Due to the high dynamic range of brightness in the real scene, the contrast degradation and detail texture distortion at different brightnesses are different, as shown in Fig. \ref{fig:Fig2}. Therefore, how to enhance the low-light images captured in wide dynamic range low-light scenes to normal exposures has received significant attention from researchers \cite{huang2022deep,li2021learning}.

In practice, one solution for this problem is to collect enormous paired low-light images from different light conditions and scenes and use these images to train an end-to-end enhancement network \cite{ma2022toward,wu2022uretinex}. However, directly using the end-to-end network often leads to under or over-exposure problems, as shown in the results of SCI \cite{ma2022toward} in Fig. \ref{fig:Fig1} (c). This is because, due to the variations in statistical and structural exposure representations under different lighting conditions, the procedures for enhancing them to normal exposures differ greatly from each other. To generate better results, researchers propose to narrow the gap between different exposure representations. For example, researchers propose to map different exposure features to the exposure-invariant feature space \cite{huang2022exposure,nsampi2021learning}. However, learning the exposure invariant space through operations such as instance normalization will inevitably disrupt the structural features of images in different scenes. These methods are also limited by the representational capacity of the model, which makes it difficult to handle multiple low-light scenes, as shown in the results of FEC \cite{huang2022deep} in Fig. \ref{fig:Fig1} (c).

In this paper, we decompose the low-light enhancement problem into a recursive enhancement task and propose a brightness-perceiving recursive enhancement method for high dynamic range low-light image enhancement. Specifically, a novel Adaptive Contrast and Texture enhancement network (ACT-Net) is proposed as the enhancement network of the recursive framework to enhance the contrast and details of low-light images simultaneously. In ACT-Net, we introduce Central Difference Convolution (CDC) \cite{yu2020searching} convolutions to perceive high-frequency gradient information and introduce the brightness adjustment branch and gradient adjustment branch to balance image contrast and detail enhancement. To avoid under or over-exposure problems, we propose a novel Brightness Perception network (BP-Net) to predict the recursive factor of the enhancement network by perceiving the brightness distribution of the image and then utilize it to control the enhancement times of ACT-Net. Finally, to train the recursive enhancement framework, a novel unsupervised training strategy is proposed to obtain pseudo-labels to train BP-Net, and joint fine-tuning ACT-Net and BP-Net to coordinate the contrast and detail enhancement of different areas of the image to obtain better visual effects. Experiments show that our method significantly outperforms the existing methods in PSNR and SSIM metrics, as shown in Fig. \ref{fig:Fig1} (a) and (b).

The contributions can be summarized as follows:

(1) Regarding the low-light image enhancement problem as an adaptive enhancement procedure, we propose a novel recursive enhancement framework for adaptive image enhancement by perceiving brightness distribution and balancing details and contrast enhancement.

(2) A novel ACT-Net is proposed to enhance image contrast and structure simultaneously, which introduces CDC to balance details and brightness enhancement. Besides, to adaptively enhance images captured under different light levels, a novel BP-Net is proposed for perceiving brightness to control the recursive enhancement times of ACT-Net.

(3) Comprehensive experiments demonstrate that, compared with the eleven low-light image enhancement and multi-exposure correction methods, our proposed method achieves state-of-the-art performance in mixed low-light datasets with high dynamic ranges.

\begin{figure*}
\setlength{\belowcaptionskip}{-0.3cm}
    \centering
    \includegraphics[width=0.98\linewidth]{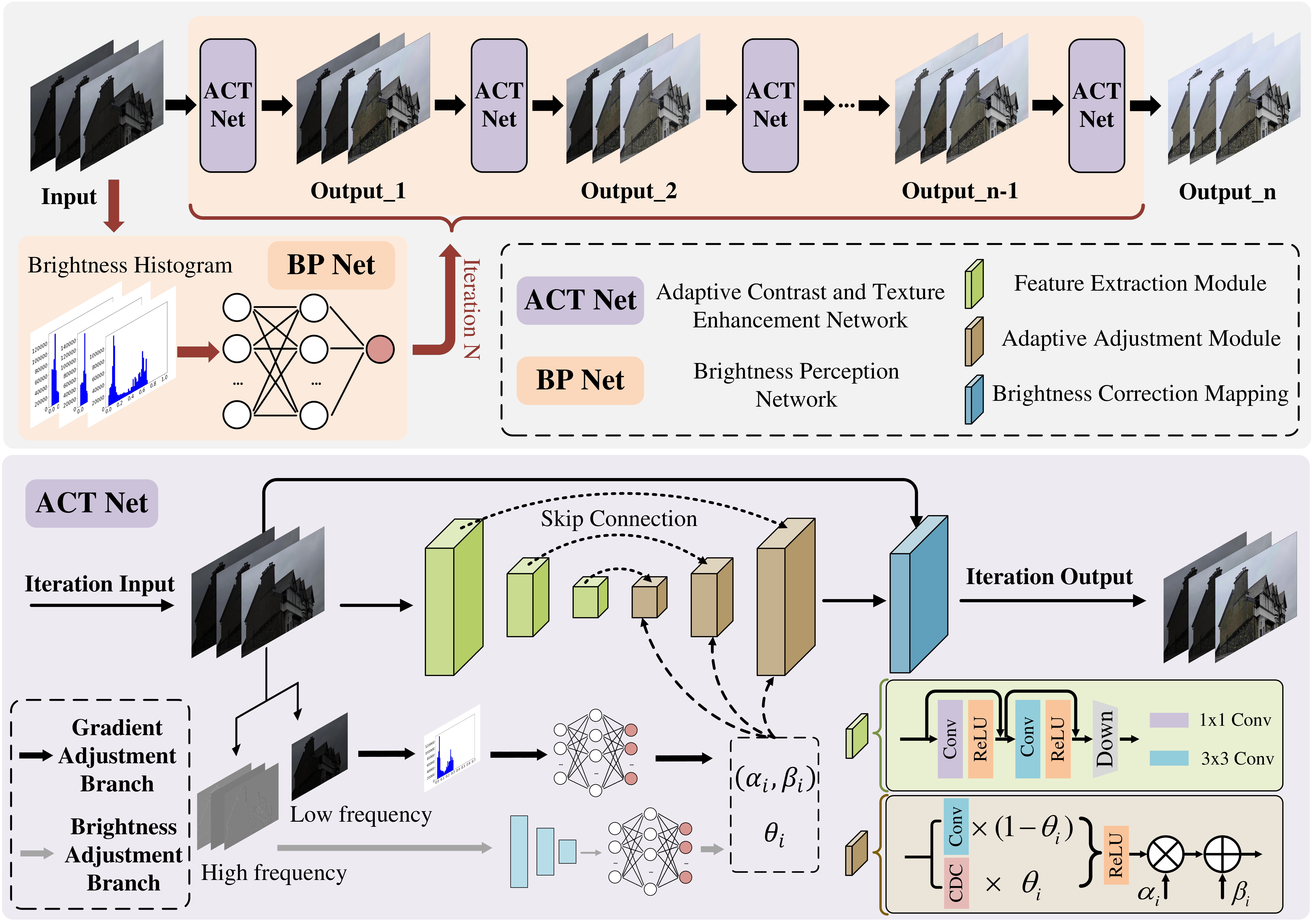}
    \caption{The overall architecture of our recursive enhancement framework, which consists of the ACT-Net and BP-Net.}
    \label{fig:Framework}
\end{figure*}
\section{Related Work}

\subsection{Low-Light Image Enhancement}

Low-Light Image Enhancement (LLIE) aims to automatically improve the visibility of images captured under low-light conditions, which can be divided into traditional methods and learning-based methods \cite{zheng2022low,liu2021benchmarking,li2021low,jiang2022degrade,jiang2022unsupervised}. Most traditional methods usually utilize Histogram Equalization \cite{pizer1987adaptive,abdullah2007dynamic,reza2004realization} and Retinex theory \cite{lee2007efficient,land1977retinex,cai2017joint,add_reference_5} to enhance image contrast and details. However, since these methods heavily rely on handcrafted image priors, they usually cannot generate satisfactory results in real-world images \cite{zheng2022low,wei2018deep}. In recent years, with the development of deep learning, many learning-based low-light image enhancement methods have been proposed, which have made significant progress \cite{wang2022local,dudhane2022burst,dong2022abandoning,wu2022uretinex,wang2022local,dudhane2022burst,zhang2022deep,zheng2022semantic}. Wei \textit{et al.} \cite{wei2018deep} propose RetinexNet to estimate the illumination map and enhance the low-light images. Wang \textit{et al.} \cite{wang2019underexposed} propose an end-to-end image-to-illumination mapping network to enhance underexposed photos. Wu \textit{et al.} \cite{wu2022uretinex} propose a Retinex-based deep unfolding network that unfolds the LLIE problem into a learnable network to decompose a low-light image into reflectance and illumination layers. However, most of the above learning-based LLIE methods rely on paired data for supervised training, which makes it hard to collect a large-scale dataset in practical environments. Therefore, unsupervised LLIE methods have attracted a lot of attention. For example, Jiang \textit{et al.} \cite{jiang2021enlightengan} propose an unsupervised EnGAN with a global-local discriminator and UNet structure generator for LLIE. Ma \textit{et al.} \cite{ma2022toward} propose an unsupervised self-calibrated illumination learning framework for efficiently enhancing low-light images. However, most existing LLIE methods can only enhance low-light images in specific low-light conditions and are difficult to adapt to real scenes with high dynamic low-light ranges.

\subsection{Multi-exposure Correction}

Multi-exposure correction (MEC) aims to learn a unified network for enhancing images captured under and over-exposure conditions \cite{afifi2021learning,huang2022exposure,huang2022deep,nsamp2018learning,huang2022exposure}. Afifi \textit{et al.} \cite{afifi2021learning} propose a pyramid structure network to enhance under and over-exposure images to normal exposure in a coarse-to-fine manner. Huang \textit{et al.} \cite{huang2022deep} propose a deep Fourier-based exposure correction network to reconstruct the representation of image lightness and structure components progressively. %Although these methods can enhance images captured by improper exposure in a high dynamic range, they cannot perceive image brightness for adaptive quality enhancement.
Although these methods can enhance the high dynamic range image caused by improper exposure to a certain level of brightness, however, due to the lack of brightness perception of the original image, they cannot further accurately control the brightness of the enhanced image to produce pleasant results.

\section{Recursive low-light Enhancement Network}

In this section, we introduce the proposed brightness perceiving recursive enhancement framework, which consists of the Adaptive Contrast and Texture enhancement network (ACT-Net) and Brightness Perception network (BP-Net), as shown in Fig. \ref{fig:Framework}. The BP-Net perceives the brightness distribution of the input low-light images and predicts the recursive factor for jumping out of the recursive process. Meanwhile, we design a novel unsupervised training strategy to facilitate the training procedure. More details are described below.

\subsection{Brightness Perception Network}
\label{sec:BPN}

In order to adaptively enhance image contrast and avoid under or over-exposure problems, we propose the brightness perception network to estimate the recursive enhancement times of ACT-Net, as shown in Fig. \ref{fig:Framework}. Firstly, in order to remove the influence of color information and reduce redundant information for brightness perception, we translate the input low-light image from RGB color space to HSV space and only reserve the brightness information in the V channel. Secondly, we calculate the histogram of this V channel to obtain brightness statistics information. Finally, BP-Net utilizes the statistics histogram as input to estimate recursive enhancement times and is composed of three FC-ReLU layers and a Sigmoid activate function. The BP-Net is trained by the following loss function:
\begin{equation}
L_{p} = \left | BP(x) - Label_{p} \right | .
\end{equation}
Where $x$ represents the input low-light image, $Label_{p}$ represents the pseudo label obtained by the unsupervised training strategy in Sec. \ref{section:Strategy}. The max and min output iterations of BP-Net are defined as $\rho_{max}$ and $\rho_{min}$, which means $BP(x)\in \left [\rho_{min}, \rho_{max}\right ]$.

\subsection{Adaptive Contrast and Texture Enhancement Network}
\label{sec:ACT-Net}

We propose a novel Adaptive Contrast and Texture enhancement network (ACT-Net) to recursively enhance the details and contrast of low-light images captured in high dynamic range environments. ACT-Net uses the UNet-style structure network as the baseline, including feature extraction modules, adaptive adjustment modules, and brightness correction modules, as shown in Fig. \ref{fig:Framework}. For feature extraction modules, we use $1 \times 1$ and $3 \times 3$ convolutional networks with ReLU to extract image features and add a downsampling layer to reduce the spatial scale. 
To effectively enhance the contrast and details of the image during the feature decoding, we designed the Adaptive Adjustment Module (AAM) to embed brightness and gradient information into the decoder, thereby performing feature modulation on the middle layer.

Specifically, we generate affine transformation parameters by extracting image brightness as prior information to modulate the middle layer features of the network and adaptively correct brightness. Additionally, we designed a brightness adjustment branch (four FC-ReLU layers) to extract the corresponding brightness modulation factors $\alpha$ and $\beta$. 
In order to enhance the image‘s texture structure while correcting brightness, we introduced Central Differential Convolution (CDC) \cite{yu2020searching} to embed gradient information, which has been proven to extract robust information independent of lighting. We also designed a gradient adjustment branch to obtain the gradient adjustment coefficient $\theta$ to automatically balance brightness and texture enhancement, which consists of three cascaded 3×3 convolutions with ReLU. 

To avoid the interference of low-frequency components, we decompose the input of each iteration into high- and low-frequency components through haar wavelet decomposition and employ the high-frequency components instead of the original image as the input of the gradient adjustment branch. Meanwhile, we extract a brightness statistical histogram from the low-frequency component of the image as input to the brightness adjustment branch, which means converting the input from the spatial domain into the statistical domain to avoid further interference of redundant information. Adaptive Adjustment Module can be expressed as the following:
\begin{equation}
\small
AAM_i(f)=ReLU(\theta_i \cdot CDC(f) + (1-\theta_i) \cdot Conv(f)) \cdot \alpha_i  + \beta_i .
\end{equation}
where $AAM_i$ represents the $i^{th}$ the adaptive adjustment modules, $f$ represents the input feature, $\theta_i$ represents the gradient adjustment coefficient, $\alpha_i$ and $\beta_i$ represent brightness adjustment factors.

Finally, we map the output of the ACT-Net to image space by brightness correction mapping \cite{li2021learning}, which can be expressed as follows:
\begin{equation}
\small
E_n(x)=E_{n-1}(x) +ACT(x) \cdot E_{n-1}(x) \cdot (1-E_{n-1}(x)).
\end{equation}
where $x$ represents input low-light images, $n$ is the number of iterations, $E_n$ represents the enhanced result of the $n^{th}$ iteration. $E_0$ represents the original input low-light image.

In order to effectively improve the contrast, suppress the color deviation, and maintain the smoothness of the image during enhancement, we use three representative loss functions for training ACT-Net in an unsupervised manner:

To control the exposure level of the enhanced image, we use exposure control loss $L_{exp}$:
\begin{equation}
L_{exp}=\frac{1}{R}\cdot  \sum_{k=1}^{R} \left \| \bar{V_k}(ACT(x)) - E \right \| .
\end{equation}
where $ACT(x)$ represents the image enhanced by ACT-Net, ${V}$ is the non-overlapping local patch, patch size of ${V}$ is $16 \times 16$, ${\bar{V_k}}$ represents the average brightness value of the $k^{th}$ local patch, ${R}$ represents the number of ${V_k}$, ${E}$ is manual exposure from \cite{li2021learning}, which is set to 0.6 \footnote{We have carried out statistical experiments on many natural images under normal exposure, and observed that most of the average brightness of these images is close to 0.6.}, and $ACT(x) \in \left [0, 1\right ] $.

To suppress the potential color deviations in the enhanced image, we use color constancy loss $L_{col}$:
\begin{equation}
% \small
L_{col}=\sum_{\forall(p, q) \in \varepsilon}\left(J^{p}-J^{q}\right)^{2}, \varepsilon=\{(R, G),(R, B),(G, B)\}.
\end{equation}
where $J^{p}$ denotes the average intensity value of $p$ channel in the enhanced image, $(p, q)$ represents a pair of channels. $R$, $G$, and $B$ represent the red, green, and blue channels in the RGB color space, respectively.

We use illumination smoothness loss $L_{t v_{m}}$ to preserve the monotonicity relations between neighboring pixels in the output of ACT-Net:

\begin{equation}
L_{t v_{m}}=\frac{1}{N} \sum_{n=1}^{N} \sum_{c \in \xi}\left(\left|\nabla_{x} m_{n}^{c}\right|+\nabla_{y} m_{n}^{c} \mid\right)^{2}, \xi=\{R, G, B\}.
\end{equation}
where $m$ represents output mapping feature of ACT-Net, $N$ is the number of iteration, $\nabla_{x}$ and $\nabla_{y}$ represent the horizontal and vertical gradient operations, respectively.

\subsection{Total Loss Function}
We use the following $L_{total} $ for training:
% The total loss can be expressed as:
\begin{equation}
L_{total} = \lambda _{exp}\cdot L_{exp} +  \lambda _{col}\cdot L_{col}  + \lambda _{tv_{m}}\cdot L_{tv_{m}}   + \lambda _{p}\cdot L_{p}.
\end{equation}

where $\lambda _{exp}$, $\lambda _{col}$, $\lambda _{tv_{m}}$ and $\lambda _{p}$ are trade-off factors. In this paper, we set $\lambda _{exp}$ to 1, $\lambda _{col}$ to 0.5, $\lambda _{tv}$ to 200, and $\lambda _{p}$ to 0.001, respectively.

\begin{figure*}
    \centering
    \setlength{\abovecaptionskip}{0.15cm}
\setlength{\belowcaptionskip}{0cm}   \includegraphics[width=1.0\linewidth]{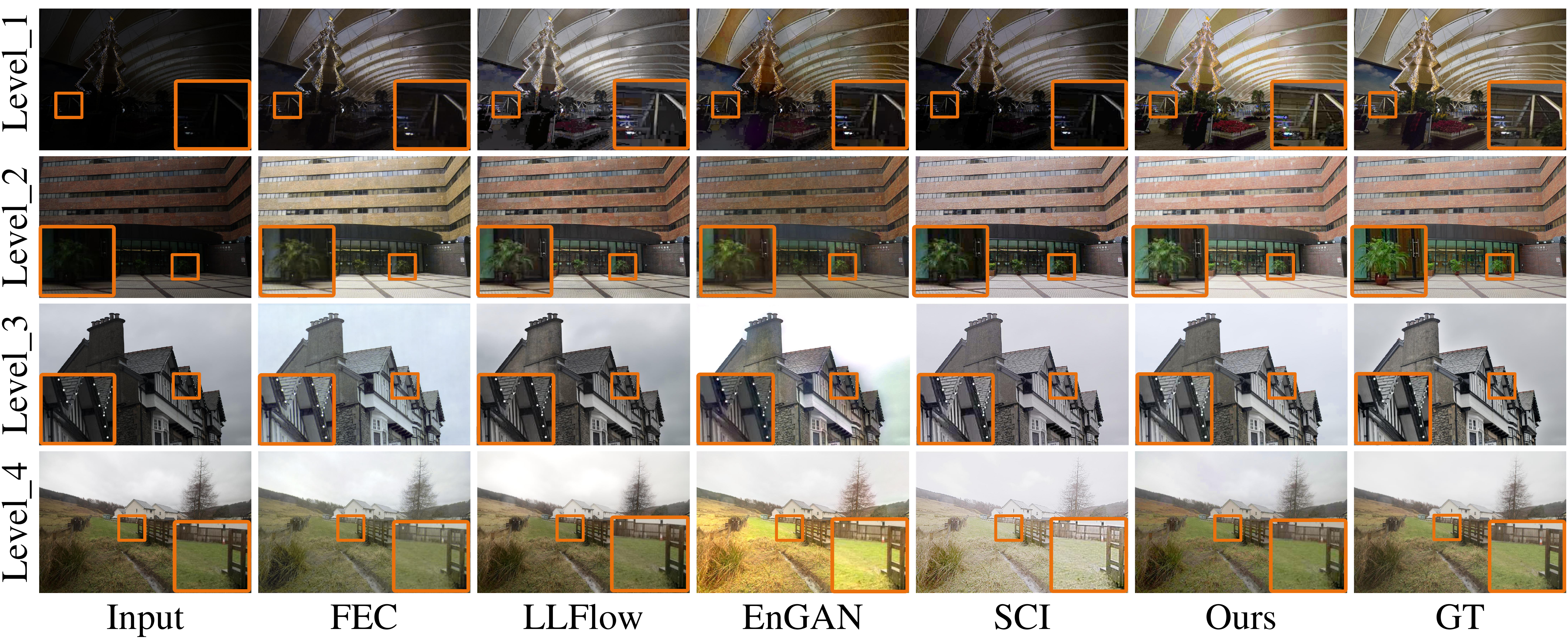}
    \caption{Qualitative comparison with existing state-of-the-art methods. From the top to bottom row are the low-light conditions from Level\_1 to Level\_4, where Level\_1 represents the darkest low-light images, and Level\_4 represents the brightest.}
    \label{fig:Qualitative_comparison}
    \vspace{-3.0mm}
\end{figure*}

\subsection{Proposed Unsupervised Training Strategy}
\label{section:Strategy}
To effectively train the proposed framework, we propose a novel unsupervised training strategy consisting of the following three steps: 

\textbf{(a) Pre-train Adaptive Contrast and Texture enhancement network (ACT-Net)}. We only use low-light images with brightness Level\_4 to train the ACT-Net, manually set the recursive enhancement times of the ACT-Net to 1, and train the ACT-Net in an unsupervised manner through the loss functions $L_{exp}$, $L_{col}$ and $L_{tv_{m}}$ in Sec. \ref{sec:ACT-Net}. After training, the ACT-Net has the ability to enhance Level\_4 low-light images to normal exposure images. 

\textbf{(b) Pre-train Brightness Perception network (BP-Net)}. We freeze the parameters of the ACT-Net to obtain $Label_{p}$, which is used to train the BP-Net. Specifically, for images from Level\_4, we set $Label_{p}$ to 1; For images from Level\_1 to Level\_3, we use ACT-Net to recursively enhance images until the average brightness of the enhanced image is approximately equal to 0.6, and then we record this recursive enhancement times as $Label_{p}$ for training BP-Net. We optimize the parameters of BP-Net by the L1 loss between BP-Net output and $Label_{p}$, as described in Sec. \ref{sec:BPN}. 

\textbf{(c) Fine-tuning overall network}. Since the manual normal exposure of 0.6 \cite{li2021learning} is a hard global threshold, it is difficult to enhance the brightness of different local image patches adaptively. Therefore, we utilize images of all brightness levels to fine-tune the overall network by $L_{total}$.

To make it easier to understand our proposed unsupervised training strategy in the manuscripts, we present a detailed pseudo-code of the proposed unsupervised training strategy, as shown in Algorithm \ref{fig:pseudocode}.

\section{Experiments and Analysis}

\subsection{Experiment setting}
\textbf{Datasets.}
We mix three prevailing low-light enhancement datasets for training and evaluation.
\begin{center}
\begin{minipage}{1\linewidth}
\centering
\begin{algorithm}[H]
\centering
	\caption{Unsupervised Training Strategy} 
	\label{train} 
	\begin{algorithmic}[]
	\REQUIRE  \begin{flushleft}low light image divided into $\left \{ Level\_1, Level\_2, Level\_3, Level\_4 \right \}$\end{flushleft}
        \STATE \textbf{\emph{Step1: pre-train ACT-Net}}\;
        
        \STATE Input: $\left \{Level\_4\right \}$
        \FOR{ image in $\left \{Level\_4\right \}$}
        \STATE optimize ACT-Net by unsupervised image enhancement loss(ACT-Net(image))
        %\STATE calculate loss(ACT-Net)
        \ENDFOR
        \STATE \textbf{\emph{Step2: pre-train BP-Net}}\;
        \STATE Input: $\left \{ Level\_1, Level\_2, Level\_3, Level\_4 \right \}$
        \STATE freeze the parameters of ACT-Net
        \FOR{ image in $\left \{ Level\_1, Level\_2, Level\_3, Level\_4 \right \}$}
        \STATE iter\_num = 0
        \WHILE{ global brightness average of image $<$ 0.6} 
        \STATE image = ACT-Net(image)
		\STATE iter\_num += 1
		\ENDWHILE
        \STATE pseudo-label = iter\_num
        \STATE \begin{flushleft}train BP-Net by calculate L1Loss(BP-Net(image),pseudo-label)
        \end{flushleft}
        \ENDFOR
        \STATE \textbf{\emph{Step3: fine-tuning overall network}}\;
        \STATE Input: $\left \{ Level\_1, Level\_2, Level\_3, Level\_4 \right \}$
        \FOR{ image in $\left \{ Level\_1, Level\_2, Level\_3, Level\_4 \right \}$}
        \STATE jointly optimizes ACT-Net and BP-Net by total loss
        \ENDFOR
	\end{algorithmic}

 \label{fig:pseudocode}
\end{algorithm}
  \vspace{-0.5mm} 
\end{minipage}
\end{center}
 Specifically, we mix low-light images from LOL \cite{wei2018deep}, SICE \cite{cai2018learning}, and MIT-Adobe FiveK \cite{bychkovsky2011learning} datasets, calculate the average brightness of each image on the V channel in HSV color space and then divide these images according to the images' average brightness range from $[0,0.15)$, $[0.15,0.3)$, $[0.3,0.45)$ to $[0.45,0.6]$ (max average brightness is lower than 0.6), denoted as Level\_1 to Level\_4, respectively. Finally, we perform a large low-light range dataset with high dynamic range, including 1167 images of Level\_1, 1382 images of Level\_2, 1130 images of Level\_3, and 889 images of Level\_4, and the remaining 84 images for testing.

To explore the generalization of the proposed method in real scenarios, we conduct further testing on five unpaired low-light image datasets, including NPE \cite{wang2013naturalness}, LIME \cite{LIME}, MEF \cite{MEF}, DICM \cite{DICM}, and VV \footnote{ https://sites.google.com/site/vonikakis/datasets}.

\textbf{Training setting.}
In our experiment, we use the Adam optimizer with a batch size of 8 and an image patch size of $512 \times 512$ to train our methods on one NVIDIA 1080Ti GPU. The learning rate is set to $0.0001$, the total epoch is set to $200$, and the $\rho_{max}$ and $\rho_{min}$ are set to $10$ and $1$.
\begin{figure*}
    \centering
    \setlength{\abovecaptionskip}{0.15cm}
\setlength{\belowcaptionskip}{0cm}   \includegraphics[width=1.0\linewidth]{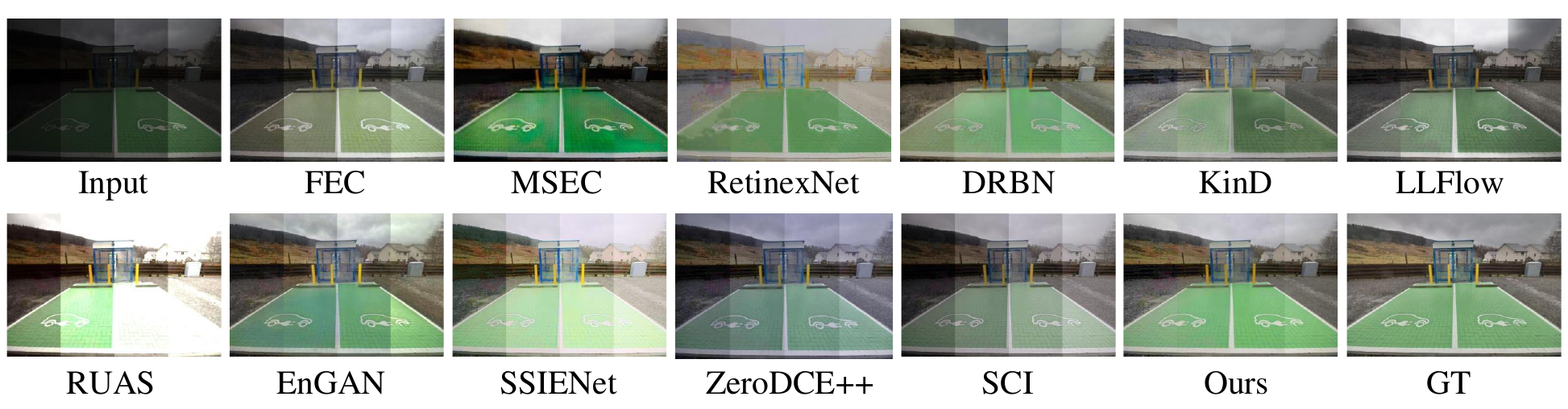}
    \caption{Comparison of enhancement results in the same scene with existing representative methods. From the left to right in each image are the low-light conditions from darkest to brightest.
    }
    \label{fig:all_method}
    \vspace{-3.0mm}
\end{figure*}
\begin{figure*}
% \vspace{-3.0mm}
    \centering
    \includegraphics[width=0.95\linewidth]{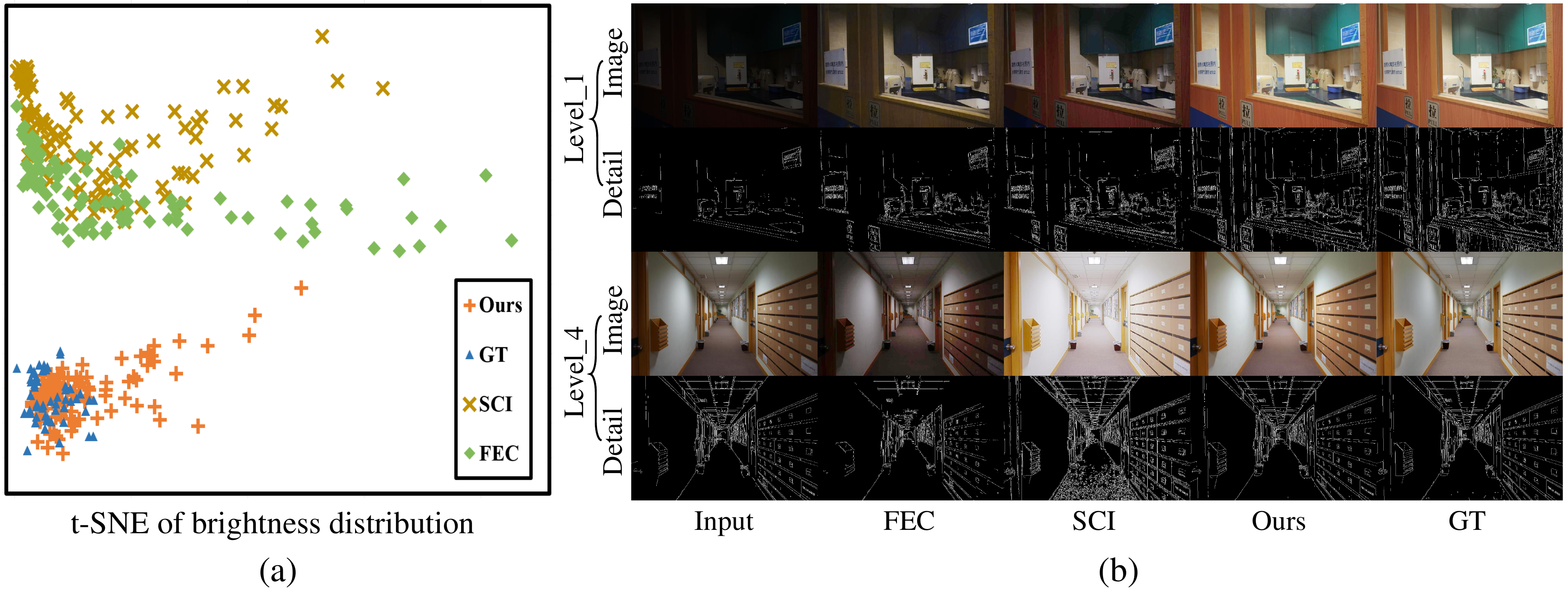}
    \caption{(a) The t-SNE visualization of brightness distribution of enhanced results. (b) The comparison of image detail enhancement result. The details are extracted by the Canny edge detector.}
    \label{fig:tsne2}
\end{figure*}
\begin{figure*}
    \centering
    \setlength{\abovecaptionskip}{0.15cm}
\setlength{\belowcaptionskip}{0cm}   \includegraphics[width=1.0\linewidth]{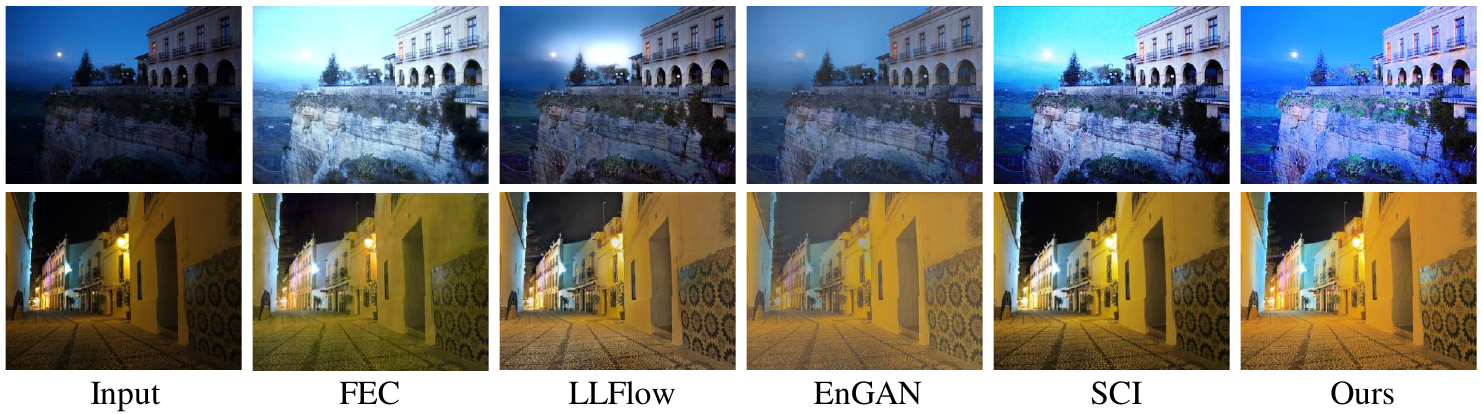}
    \caption{Comparison with existing representative methods on five representative real-world datasets (including NPE, LIME, MEF, DICM, and VV).}
    \label{fig:VV_comparison}
    \vspace{-3.0mm}
\end{figure*}

\textbf{Comparison methods.}
We compare the proposed method with five state-of-the-art unsupervised LLIE methods (including SCI \cite{ma2022toward}, SSIENet \cite{zhang2020self}, ZeroDCE++ \cite{li2021learning}, RUAS \cite{liu2021retinex} EnGAN \cite{jiang2021enlightengan}), four advanced supervised learning low-light image enhancement methods (including RetinexNet \cite{wei2018deep}, KinD \cite{zhang2021beyond}, DRBN \cite{yang2020fidelity}, LLFlow \cite{wang2022low}) and two MEC methods (including MSEC \cite{afifi2021learning} and FEC \cite{huang2022deep}).

\textbf{Metrics.} To comprehensively demonstrate the superiority of our method, we use three reference metrics and three no-reference metrics to evaluate the performance. For reference metrics, we use the PSNR $\uparrow$\footnote{ $\uparrow$ means the higher, the better, $\downarrow$ means the lower, the better.} \cite{huynh2008scope}, SSIM $\uparrow$ \cite{wang2004image} and LPIPS $\downarrow$\footnotemark[2] \cite{zhang2018unreasonable}. For no-reference metrics, we use NIQE $\downarrow$ \cite{mittal2012making}, EME $\uparrow$ \cite{agaian2000new} and LOE $\downarrow$ \cite{wang2013naturalness} for evaluation.

\begin{table*}[ht]
\small
\centering
\caption{The PSNR$\uparrow$ and SSIM$\uparrow$ results for eleven comparison methods across four different brightness levels on our constructed datasets (including LOL \cite{wei2018deep}, SICE \cite{cai2018learning}, and MIT-Adobe FiveK \cite{bychkovsky2011learning}). The best and second-best performances are marked as {\color{red} \textbf{bold}} and {\color{blue} \textbf{bold}}, respectively.
}

\begin{tabular}{c|llllllllcc}
\hline
\Xhline{2.\arrayrulewidth}
                          & \multicolumn{2}{c}{Level\_1}                                                   & \multicolumn{2}{c}{Level\_2}                                                   & \multicolumn{2}{c}{Level\_3}                                                   & \multicolumn{2}{c}{Level\_4}                                                   & \multicolumn{2}{c}{Average}                                                    \\ \cline{2-11} 
\multirow{-2}{*}{Methods} & \multicolumn{1}{c}{PSNR}               & \multicolumn{1}{c}{SSIM}              & \multicolumn{1}{c}{PSNR}               & \multicolumn{1}{c}{SSIM}              & \multicolumn{1}{c}{PSNR}               & \multicolumn{1}{c}{SSIM}              & \multicolumn{1}{c}{PSNR}               & \multicolumn{1}{c}{SSIM}              & PSNR                                   & SSIM                                  \\ \hline
MSEC                      & 17.056                                 & 0.671                                 & 18.736                                 & 0.689                                 & {\color{blue} \textbf{19.291}} & 0.771                                 & 18.826                                 & 0.724                                 & {\color{blue} \textbf{18.477}} & 0.714                                 \\
FEC                       & 17.791                                 & 0.685                                 & 18.938                                 & 0.767                                 & 18.166                                 & 0.738                                 & {\color{blue} \textbf{18.867}} & {\color{red} \textbf{0.842}} & 18.440                                 & 0.758                                 \\
RetinexNet                & {\color{blue} \textbf{18.362}} & 0.731                                 & 18.524                                 & 0.776                                 & 16.813                                 & 0.766                                 & 14.865                                 & 0.760                                 & 17.141                                 & 0.758                                 \\
DRBN                      & 16.161                                 & 0.658                                 & 16.385                                 & 0.683                                 & 16.592                                 & 0.713                                 & 16.851                                 & 0.734                                 & 16.498                                 & 0.697                                 \\
KinD                      & 17.276                                 & 0.715                                 & 17.881                                 & 0.790                                 & 18.726                                 & 0.795                                 & 17.292                                 & 0.756                                 & 17.794                                 & 0.764                                 \\
LLFlow                    & 17.482                                 & 0.723                                 & 16.969                                 & 0.770                                 & 17.666                                 & {\color{blue} \textbf{0.803}} & 16.867                                 & 0.806                                 & 17.246                                 & {\color{blue} \textbf{0.775}} \\
EnGAN                     & 16.488                                 & 0.704                                 & 18.907                                 & {\color{blue} \textbf{0.812}} & 17.891                                 & 0.793                                 & 17.482                                 & 0.772                                 & 17.692                                 & 0.771                                 \\
RUAS                      & 13.919                                 & 0.661                                 & 14.581                                 & 0.676                                 & 10.974                                 & 0.642                                 & 9.790                                  & 0.543                                 & 12.316                                 & 0.630                                 \\
SSIENet                   & 17.042                                 & {\color{red} \textbf{0.781}} & 15.989                                 & 0.806                                 & 13.898                                 & 0.755                                 & 10.969                                 & 0.728                                 & 14.475                                 & 0.768                                 \\
ZeroDCE++                 & 16.099                                 & 0.683                                 & 16.721                                 & 0.759                                 & 17.035                                 & 0.791                                 & 17.121                                 & 0.794                                 & 16.744                                 & 0.757                                 \\
SCI                       & 16.883                                 & 0.655                                 & {\color{blue} \textbf{18.787}} & 0.710                                 & 19.133                                 & 0.796                                 & 18.381                                 & 0.734                                 & 18.296                                 & 0.724                                 \\
Ours                      & {\color{red} \textbf{19.065}} & {\color{blue} \textbf{0.752}} & {\color{red} \textbf{19.702}} & {\color{red} \textbf{0.819}} & {\color{red} \textbf{19.435}} & {\color{red} \textbf{0.813}} & {\color{red} \textbf{19.193}} & {\color{blue} \textbf{0.822}} & {\color{red} \textbf{19.349}} & {\color{red} \textbf{0.802}} \\ \hline
\Xhline{2.\arrayrulewidth}
\end{tabular}

\label{tab:performance_each_level}
\end{table*}
\begin{table*}[ht]
\small
\centering
\caption{Comparison with nine existing low light enhancement methods on our constructed datasets by three no-reference and three reference metrics. The best and second-best performances are marked as {\color{red} \textbf{bold}} and {\color{blue} \textbf{bold}}, respectively.}

\begin{tabular}{c|cccccccccc}
\hline
\Xhline{2.\arrayrulewidth}
                          & \multicolumn{4}{c|}{Supervised Method}                                                                             & \multicolumn{6}{c}{Unsupervised Method}                                                                                                                                     \\ \cline{2-11} 
\multirow{-2}{*}{Metrics} & \multicolumn{1}{c}{RetinexNet} & \multicolumn{1}{c}{DRBN} & \multicolumn{1}{c}{KinD} & \multicolumn{1}{c|}{LLFlow} & \multicolumn{1}{c}{EnGAN}   & \multicolumn{1}{c}{RUAS} & \multicolumn{1}{c}{SSIENet} & \multicolumn{1}{c}{ZeroDCE++} & \multicolumn{1}{c}{SCI}       & \multicolumn{1}{c}{Ours}      \\ \hline
PSNR↑             & 17.14                          & 16.50                    & 17.79                    &  \multicolumn{1}{c|}{17.25}                      & 17.69                       & 12.32                    & 14.47                       & 16.74                         & {\color{blue} \textbf{18.30}}                        & {\color{red} \textbf{19.34}}  \\
SSIM↑                     & 0.76                           & 0.70                     & 0.76                     &  \multicolumn{1}{c|}{{\color{blue} \textbf{0.78}}}  & 0.77                        & 0.63                     & 0.77                        & 0.76                          & 0.72                          & {\color{red} \textbf{0.80}}   \\
LPIPS↓                    & 0.20                           & 0.24                     & 0.18                     & \multicolumn{1}{c|}{{\color{blue}\textbf{0.16}}} & 0.17                        & 0.22                     & 0.18                        & 0.16                          & 0.18                          & {\color{red} \textbf{0.14}}   \\
LOE↓                      & 745.99                         & 555.75                   & 448.32                   & \multicolumn{1}{c|}{458.45}                      & 566.22                      & 576.19                   & 422.22                      & 448.11                        & {\color{blue} \textbf{408.81}} & {\color{red} \textbf{398.94}} \\
NIQE↓                     & 2.65                           & 2.92                     & 2.53                     & \multicolumn{1}{c|}{{\color{blue} \textbf{2.41}}} & 2.77                        & 4.61                     & 3.05                        & 2.41                          & 2.49                          & {\color{red}\textbf{ 2.39}}   \\
EME↑                      & 4.17                           & 5.17                     & 5.09                     & \multicolumn{1}{c|}{5.49}                        & {\color{red} \textbf{5.86}} & 4.73                     & 4.59                        & 5.13                          & 5.25                          & {\color{blue} \textbf{5.68}}   \\ \hline
\Xhline{2.\arrayrulewidth}
\end{tabular}
\label{tab:performance_each_metric}
\end{table*}

\begin{table*}[ht]
\small
\centering
\caption{Comparison of the average performance of existing eleven methods on five representative real-world datasets (including NPE, LIME, MEF, DICM, and VV) by three no-reference metrics. The best performance and second best performance are marked as {\color[HTML]{FF0000} \textbf{bold}} and {\color{blue} \textbf{bold}}, respectively.}
\resizebox{\textwidth}{!}{
\begin{tabular}{c|cccccc|cccccc}
\Xhline{2.\arrayrulewidth}
\hline
                          & \multicolumn{6}{c|}{Supervised Method}                                       & \multicolumn{6}{c}{Unsupervised Method}                                                                                    \\ \cline{2-13} 
\multirow{-2}{*}{Metrics} & FEC    & \multicolumn{1}{c|}{MSEC}   & RetinexNet & DRBN   & KinD   & LLFlow & EnGAN                       & RUAS   & SSIENet & ZeroDCE++ & SCI                           & Ours                          \\ \hline
LOE↓                      & 479.57 & \multicolumn{1}{c|}{581.05} & 793.68     & 673.60 & 464.32 & 431.48 & 549.43                      & 541.71 & 407.96  & 385.37    & {\color{blue} \textbf{376.92}} & {\color{red} \textbf{363.58}} \\
NIQE↓                     & 3.52   & \multicolumn{1}{c|}{3.75}   & 3.91       & 3.92   & 3.72   & 3.69   & {\color{blue} \textbf{3.42}} & 4.96   & 4.40    & 3.53      & 3.84                          & {\color{red} \textbf{3.40}}   \\
EME↑                      & 5.77   & \multicolumn{1}{c|}{6.18}   & 3.78       & 5.99   & 6.57   & 6.73   & {\color{red} \textbf{6.81}} & 5.68   & 5.12    & 6.49      & 6.63                          & {\color{blue} \textbf{6.77}}   \\ \hline
\Xhline{2.\arrayrulewidth}
\end{tabular}
}
\label{tab:VV_compare}
\end{table*}
\begin{table}[ht]
\small
\centering
% \small
\caption{ Comparison with two multi-exposure correction methods on our constructed datasets by three no-reference and three reference metrics. The best performance is marked as {\color{red} \textbf{bold}}.
}
\setlength{\tabcolsep}{0.8mm}{
\begin{tabular}{c|cccccc}
\Xhline{2.\arrayrulewidth}
\hline

 & PSNR↑                                 & SSIM↑                                & LPIPS↓                               & LOE↓                                   & NIQE↓                                & EME↑                                 \\
\hline MSEC                          & 18.48                                 & 0.71                                 & 0.20                                 & 525.72                                 & 2.48                                 & 5.64                                 \\
FEC                           & 18.44                                 & 0.76                                 & 0.18                                 & 618.79                                 & 2.68                                 & 4.56                                 \\
Ours                           & {\color{red} \textbf{19.35}} & {\color{red} \textbf{0.80}} & {\color{red} \textbf{0.15}} & {\color{red} \textbf{398.94}} & {\color{red} \textbf{2.39}} & {\color{red} \textbf{5.68}} \\ \hline
\Xhline{2.\arrayrulewidth}
\end{tabular}}
\label{tab:performance_mec}
%\vspace{-3.0mm}
\end{table}

\subsection{Qualitative results}
Fig. \ref{fig:Qualitative_comparison} shows the qualitative comparison with existing SOTA methods in different low-light conditions from Level\_1 to Level\_4. We can observe that in low-light scenes with high dynamic ranges, images enhanced by our proposed method are superior to the existing methods in image detail and contrast. For example, in the darkest scene, as shown in the top row in Fig. \ref{fig:Qualitative_comparison}, the existing SOTA methods are unable to effectively enhance the contrast of the image, and the details of the enhanced image are lost, while our method can enhance the low-light image to the same brightness distribution as Ground Truth (GT), and also enhance the local texture details of the image, as shown in the red box. This demonstrates the effectiveness of our proposed ACT-Net. Besides, we can observe that in the brightest scene Level\_4, images enhanced by EnGAN \cite{jiang2021enlightengan} and LLFlow \cite{wang2022low} appear color distortion, and images enhanced by SCI \cite{ma2022toward} appears over-exposure problem causing the image to be overexposed, as shown in the bottom row in Fig. \ref{fig:Qualitative_comparison}. However, our proposed method can utilize BP-Net to perceive image brightness and then assist ACT-Net to adaptively enhance the low-light image to the normal brightness range, which improves image details and contrast at the same time without color distortion and over-exposure. It demonstrates that the proposed method can effectively adapt to natural scenes with varying brightness levels without additional fine-tuning.

To intuitively show the effectiveness of the proposed method, Fig. \ref{fig:all_method} showcases the enhancement results of various methods on the same scene under different lighting conditions. It can be observed that most methods exhibit under or over-exposure when dealing with different lighting scenarios. In comparison, the proposed method enhances the image more uniformly and closely to the ground truth.

To further demonstrate that the proposed method can adaptively enhance image contrast, as shown in Fig. \ref{fig:tsne2} (a), we perform the brightness histogram t-SNE visualization of the enhanced image. We can observe that the images enhanced by SCI and FEC are still under-exposure/over-exposure, causing brightness distribution to be discrete. However, the brightness distribution of images enhanced by our method is clustered and closer to GT. To demonstrate the superiority of the proposed ACT-Net in image detail enhancement, we present the image detail enhancement results in low-light Level\_1 and Level\_4 scenes, as shown in Fig. \ref{fig:tsne2} (b). It can be easily observed that the detail texture enhanced by our proposed method is closer to GT.

Furthermore, to evaluate the generalization ability of the proposed method in real night scenes, we directly evaluate our method on five real datasets (including NPE \cite{wang2013naturalness}, LIME \cite{LIME}, MEF \cite{MEF}, DICM \cite{DICM}, and VV). Note that all comparison methods are retrained on the same datasets as ours and directly tested on five other datasets without fine-tuning. As shown in Fig. \ref{fig:VV_comparison}, images enhanced by FEC \cite{huang2022deep} appear over-exposure caused by over-enhancement, and the images enhanced by LLFlow \cite{wang2022low} exhibit unnatural light spots in some scenes. In contrast, our proposed method enhances low-light images more visually pleasuring.

\begin{figure}
    \centering
    \includegraphics[width=1\linewidth]{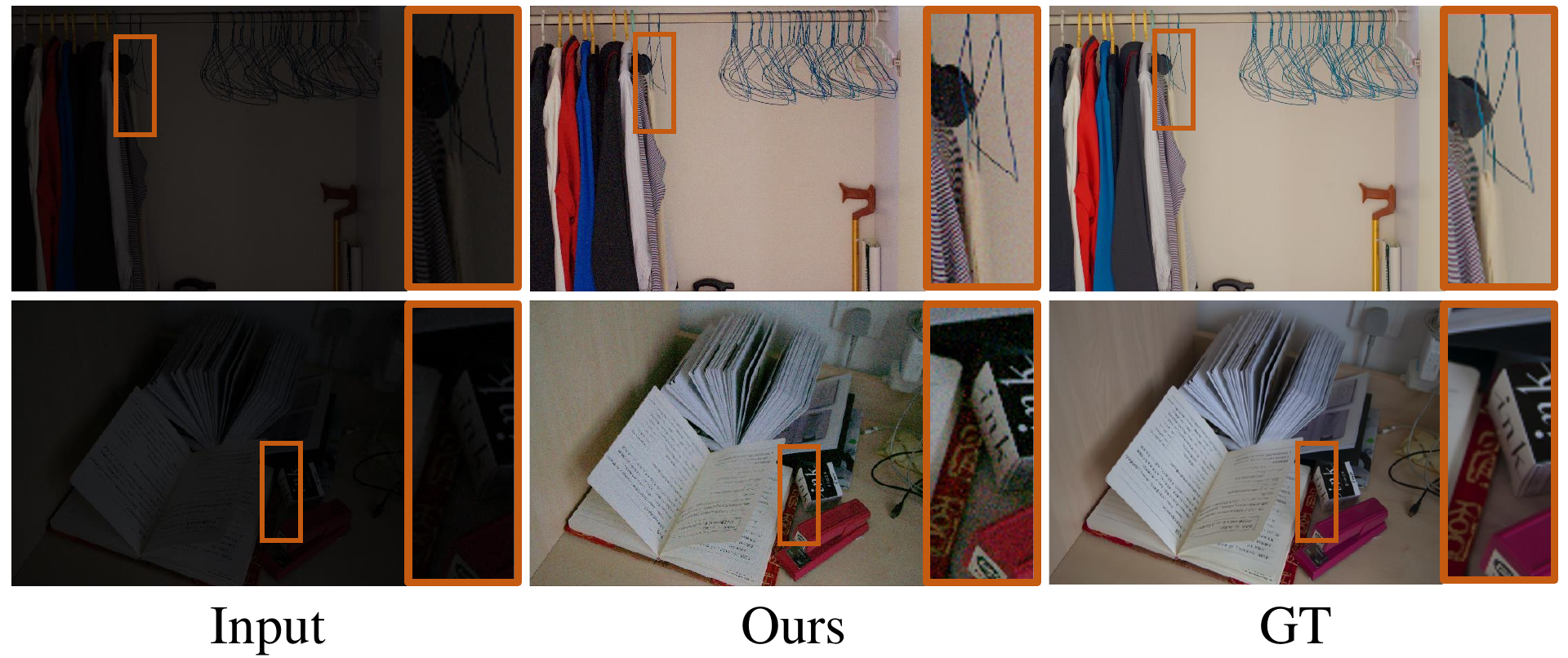}
    \caption{Examples of failed results with noise residue.}
    \label{fig:ours_false}
    \vspace{-5.0mm}
\end{figure}

\begin{table*}[ht]
\small
\centering
\caption{Comparison with existing methods on computational cost. The best performance and second best performance are marked as {\color[HTML]{FF0000} \textbf{bold}} and {\color{blue} \textbf{bold}}, respectively.}
\setlength{\tabcolsep}{3.5mm}{
\begin{tabular}{c|ccccc}
\hline
\Xhline{2.\arrayrulewidth}
        Methods                & FLOPs(G) & Params(M) & Time(S)  & PSNR↑  & SSIM \cite{wang2004image}↑\\ \hline
        \Xhline{2.\arrayrulewidth}
\multicolumn{1}{c|}{SCI \cite{ma2022toward}}    & 0.0620   & 0.000348  & 0.001688 & 18.30 & 0.72 \\
\multicolumn{1}{c|}{ZeroDCE++ \cite{li2021learning}}    & 5.2112   & 0.078912  & 0.002375 & 16.74 & 0.76 \\
\multicolumn{1}{c|}{RUAS \cite{liu2021retinex}}   & 0.2813   & 0.001437  & 0.042659 & 12.32 & 0.63 \\
\multicolumn{1}{c|}{DRBN \cite{yang2020fidelity}}   & 37.7902  & 0.577168  & 0.065768 & 16.50 & 0.70 \\
\multicolumn{1}{l|}{KinD \cite{zhang2021beyond}}   & 29.1303  & 8.540217  & 0.135406 & 17.79 & 0.76 \\
\multicolumn{1}{c|}{EnGAN \cite{jiang2021enlightengan}}  & 61.0102  & 8.636037  & 0.135406 & 17.69 & 0.77 \\
\multicolumn{1}{c|}{SSIE \cite{zhang2020self}}   & 34.6070  & 0.682400  & 0.150337 & 14.47 & 0.77 \\
\multicolumn{1}{c|}{RetinexNet \cite{wei2018deep}}     & 136.0151 & 0.838364  & 0.346362 & 17.14 & 0.76 \\
\multicolumn{1}{c|}{FEC \cite{huang2022deep}}    & 5.8156   & 0.151910  & 0.601818 & {\color{blue} \textbf{18.44}} & 0.76 \\
\multicolumn{1}{c|}{LLFlow  \cite{wang2022low}} & 26.3169  & 38.86652 & 0.717524 & 17.25 & {\color{blue} \textbf{0.78}} \\
\multicolumn{1}{c|}{Ours}    & 3.6633   & 1.140580  & 0.179724 & {\color[HTML]{FF0000} \textbf{19.35}} & {\color[HTML]{FF0000} \textbf{0.80}} \\ \hline
\Xhline{2.\arrayrulewidth}
\end{tabular}}
\label{tab:Computational Cost Comparison}
\end{table*}

Nevertheless, our method also has some limitations in dealing with the noise in real low-light images. As shown in Fig \ref{fig:ours_false}, our enhanced results have a noise residual due to the absence of a denoising module in our proposed method. And it is a crucial direction for improvement in our future work.

\subsection{Quantitative results}

In Table \ref{tab:performance_each_level}, our proposed method is compared with eleven SOTA methods on PSNR/SSIM metrics at different brightness levels. One can observe that our proposed method achieves the best performance in the majority of low-light environments. For example, compared with the existing supervised and unsupervised methods, our method outperforms the PSNR and SSIM performances by 0.915 and 0.007 under the Level\_2 low-light environment. This is because our adaptive recursive framework can perceive image brightness and perform accurate enhancement without under or over-exposure. Besides, compared with the existing unsupervised SOTA method SCI \cite{ma2022toward}, our proposed unsupervised method outperforms SCI by a large margin under all low-light environments. This demonstrates that the proposed method can be better applied to the high dynamic range low-light environment in real scenes.

To comprehensively compare the quality of enhanced images, we perform image quality comparisons on three reference metrics (PSNR↑, SSIM↑, LPIPS↓) and three no-reference metrics (LOE↓, NIQE↓, EME↑) in all brightness environments, as shown in Table \ref{tab:performance_each_metric} and Table \ref{tab:performance_mec}. One can observe that for the average performance in all brightness environments, we achieve SOTA performance on the five image quality metrics, including PSNR, SSIM, LPIPS, LOE, and NIQE, and outperform existing LLIE methods by 1.04 on PSNR, 0.02 on SSIM, 0.02 on LPIPS, 9.87 on LOE, and 0.02 on NIQE. We also outperform two SOTA multi-exposure correction methods, as shown in Table \ref{tab:performance_mec}. This demonstrates that the images enhanced by our proposed method are more natural and high-quality. 

As shown in Table \ref{tab:VV_compare}, we can observe that our method outperforms existing methods in most cases on real-world datasets. Specifically, we achieve state-of-the-art performance on two no-reference image quality metrics, outperforming the existing LLIE methods by 13.34 on LOE and 0.02 on NIQE, respectively.

\subsection{Computational cost comparison}
In order to better demonstrate the superiority of the proposed method, we compare the computational cost with the existing method, and the results are shown in Table \ref{tab:Computational Cost Comparison}. We can observe that our proposed method not only achieves state-of-the-art performance but also has a great advantage in the computational cost, compared with some large low-light image enhancement models such as LLFlow \cite{wang2022low} and FEC \cite{huang2022deep}. 

\begin{table}[ht]
\footnotesize
\caption{Ablution study results. $\surd $ and × represent with and without, respectively. }

\centering
\setlength{\tabcolsep}{1mm}{
\begin{tabular}{c|cccccc|cc}
\hline
\Xhline{2.\arrayrulewidth}
\multirow{2}{*}{}    & \multicolumn{6}{c|}{ablation study setting}                                    & \multicolumn{2}{c}{performance}  \\ \cline{2-9} 
                     & w/RF & \multicolumn{1}{c|}{w/BPN}      & w/BAB     & w/FDD & w/BH & w/GAB & PSNR↑            & NIQE↓           \\ \hline
\multirow{3}{*}{(a)} & ×     & \multicolumn{1}{c|}{×}          & ×          & ×      & ×     & ×      & 17.285          & 4.253          \\
                     & $\surd $     & \multicolumn{1}{c|}{×}          & ×          & ×      & ×     & ×      & 18.441          & 3.882          \\
                     & $\surd $     & \multicolumn{1}{c|}{\textbf{$\surd $}} & \textbf{×} & ×      & ×     & ×      & 18.771          & 2.613          \\ \hline
\multirow{4}{*}{(b)} & $\surd $     & \multicolumn{1}{c|}{$\surd $}          & $\surd $          & ×      & ×     & ×      & 18.916          & 2.578          \\
                     & $\surd $     & \multicolumn{1}{c|}{$\surd $}          & $\surd $          & $\surd $      & ×     & ×      & 19.091          & 2.527          \\
                     & $\surd $     & \multicolumn{1}{c|}{$\surd $}          & $\surd $          & $\surd $      & $\surd $     & ×      & 19.243          & 2.488          \\
                     & $\surd $     & \multicolumn{1}{c|}{$\surd $}          & $\surd $          & $\surd $      & $\surd $     & $\surd $      & \textbf{19.349} & \textbf{2.389} \\ \hline
\Xhline{2.\arrayrulewidth}
\end{tabular}}
\label{tab:ablution1}
\vspace{-4mm}
\end{table}
\begin{figure*}
    \centering
    \includegraphics[width=1\linewidth]{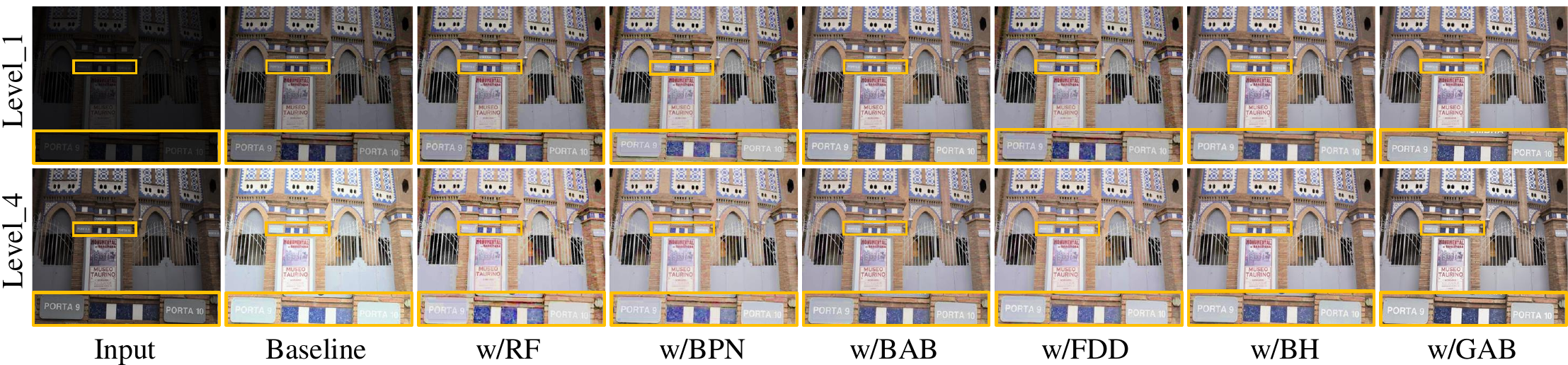}
    \caption{Visual comparison of ablation studies. From left to right are the baseline and results after add recursive framework (RF), brightness perception network (BPN), brightness adjustment branch (BAB), frequency domain decomposition (FDD), brightness histogram(BH), and gradient adjustment branch (GAB) sequentially on the basis of the baseline model.}
    \label{fig:Ablation}
    \vspace{-3.0mm}
\end{figure*}
\subsection{Ablation study}
To demonstrate the effectiveness of the proposed method, we conduct comprehensive ablation studies:

\textbf{(a) proposed framework}: recursive framework (\textbf{RF}), brightness perception network (\textbf{BPN}). We conduct ablation studies on RF and BPN, as shown in Table \ref{tab:ablution1} (a). Note that the first row in Table \ref{tab:ablution1} means baseline (\emph{i.e.,} UNet). After using RF, the input low-light images are enhanced by baseline until the average brightness is above 0.6, which improves the performance of the baseline by 1.16 and 0.371 on the PSNR and NIQE. After using BPN, the baseline can adaptively enhance input low-light images, avoiding under or over-exposure problems, which further improves the performance of the baseline by 0.33 and 1.269 on the PSNR and NIQE. As shown in Figure \ref{fig:Ablation}, compared with the baseline, w/RF and w/BPN can effectively avoid the overexposure problem, demonstrating that proposed RF and BPN can improve the quality of the enhanced image. 

\begin{figure}
    \centering
    \includegraphics[width=1\linewidth]{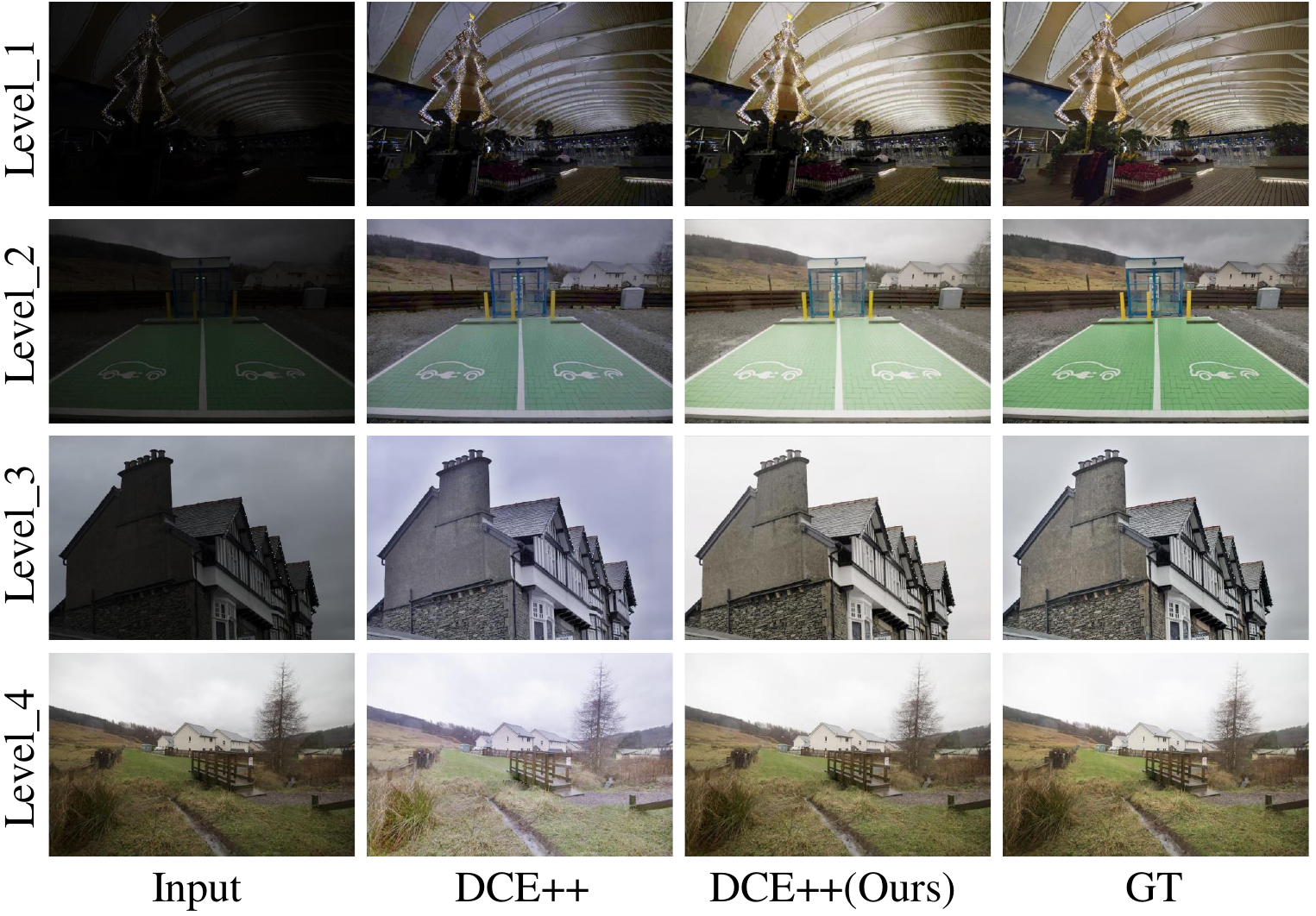}
    \caption{Visual comparison on DCE++ and DCE++ (Ours).}
    \label{fig:DCE(ours)}
    \vspace{-6.0mm}
\end{figure}

To verify the generality of the proposed framework, we replaced the backbone of our method with ZeroDCE++ \cite{li2021learning}, which is also a classical unsupervised training backbone, denoted as DCE++(ours). As shown in Fig. \ref{fig:DCE(ours)}, we can observe that compared with DCE++, the contrast enhancement and detail enhancement capabilities of DCE++(ours) have been significantly improved in different brightness scenes. It demonstrates that our proposed framework can be effectively generalized to other low-light enhancement backbones and significantly improve their performance.

\textbf{(b) Adaptive Contrast and Texture enhancement network setting}: brightness adjustment branch (\textbf{BAB}), frequency domain decomposition (\textbf{FDD}), brightness histogram (\textbf{BH}), and gradient adjustment branch (\textbf{GAB}). As can be seen in the fourth to seventh rows of Table \ref{tab:ablution1}, after using the BAB, the internal features of the adaptive adjustment decoder can perceive the brightness distribution and then more adaptively enhance the contrast of the image, which further improves the baseline performance by 0.330 on the PSNR. After equipping FDD, the performance of the baseline is further improved by 0.175 on PSNR. After equipping BH, the baseline network can perceive the image brightness distribution more efficiently, which further improves the baseline performance by 0.152 on PSNR. After equipping the GAB, the baseline can adaptively adjust the distribution of smoothness and detail features in different regions and improve the detailed edge of the low-light image, which further improves the baseline performance by 0.106 on the PSNR. We also perform the visualization results of the ablation studies on two low-light conditions, as shown in Fig. \ref{fig:Ablation}. After equipping the proposed framework and module sequentially, the contrast and details of the low-light images are gradually enhanced. It demonstrates the effectiveness of our proposed method.

\begin{table}[t]
\centering
\caption{Ablution study results for the loss function, the best performance is marked as {\color[HTML]{FF0000} \textbf{bold}}.}
\setlength{\tabcolsep}{0.5mm}{
\begin{tabular}{c|cccccc}
\hline
\Xhline{2.\arrayrulewidth}
w/   & PSNR↑ & SSIM↑ & LPIP↓ & LOE↓   & NIQE↓ & EME↑ \\ \hline
\Xhline{2.\arrayrulewidth}
$L_{p} + L_{tv}$   & 12.67 & 0.66  & 0.31   & 613.72 & 2.82  & 4.71 \\
$L_{p} + L_{exp}$    & 17.32 & 0.71  & 0.23   & 525.13 & 2.65  & 4.09 \\
$L_{p} + L_{exp} + L_{color}$ & 18.19 & 0.75  & 0.21   & 422.37 & 2.57  & 5.43 \\
$L_{p} + L_{tv} + L_{exp} + L_{color}$  & {\color[HTML]{FF0000} \textbf{19.35}} & {\color[HTML]{FF0000} \textbf{0.80}}  & {\color[HTML]{FF0000} \textbf{0.15}}   & {\color[HTML]{FF0000} \textbf{398.94}} & {\color[HTML]{FF0000} \textbf{2.39}}  & {\color[HTML]{FF0000} \textbf{5.68}} \\ \hline
\Xhline{2.\arrayrulewidth}
\end{tabular}}
\label{tab:Ablation Study For loss}
\end{table}
\begin{figure}
    \centering
    \includegraphics[width=1\linewidth]{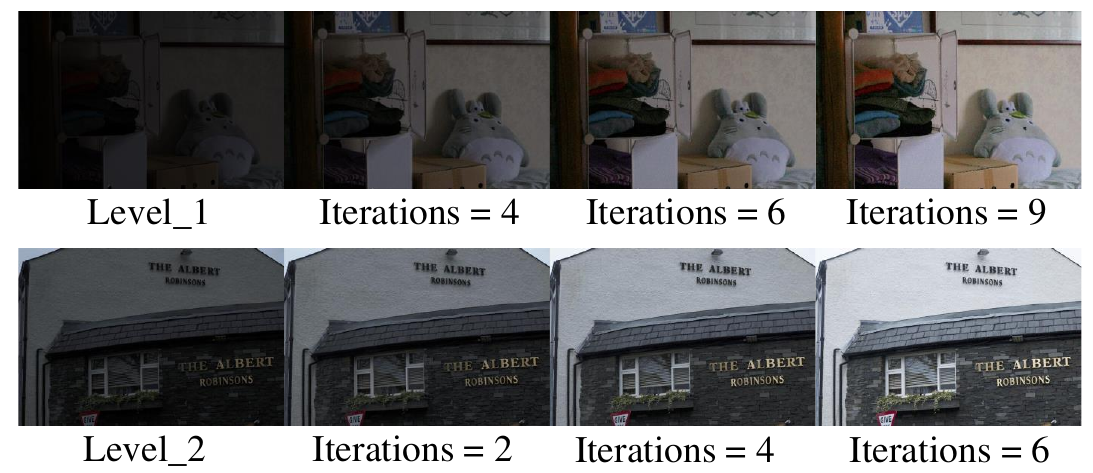}
    \caption{The visualization of intermediate enhancement results under Level\_1 and  Level\_2 low light environments.}
    \label{fig:Iteraton}
    \vspace{-5.0mm}
\end{figure}
To demonstrate the effectiveness of the proposed recursive framework, we show the intermediate enhancement results in the recursive enhancement phase under two different low-light scenes, as shown in Fig. \ref{fig:Iteraton}. One can observe that the recursive enhancement times of Level\_1 image predicted by BP-Net are more than those of the Level\_2 image. With the recursive enhancement of ACT-Net, the brightness distribution of the enhanced image is continuously improved until it reaches normal exposure. It shows that ACT-Net can recursively enhance image detail and contrast, and BP-Net can perceive the brightness distribution to guide ACT-Net to perform the adaptive enhancement.

\textbf{(c) Loss Function}:
To explore the impact of loss functions used in the manuscript, we perform an ablation study on loss functions. Since the network cannot be trained without $L_{p}$, this section conducts ablation experiments for illumination smoothness loss $L_{tv}$, exposure control loss $L_{exp}$, and color constancy loss $L_{color}$. We provide more numerical results by applying different loss functions for training. As shown in Table. \ref{tab:Ablation Study For loss}, it is hard to effectively enhance the low-light image by only utilizing $L_{p}$ and $L_{tv}$ for training, which could cause all indicators to be significantly worse than using all losses. When utilizing $L_{p}$ and $L_{exp}$, the PSNR and SSIM performances improve by 4.65 and 0.05. After applying all loss functions simultaneously, the model achieves the best performance.

\begin{figure}
    \centering
    \includegraphics[width=1\linewidth]{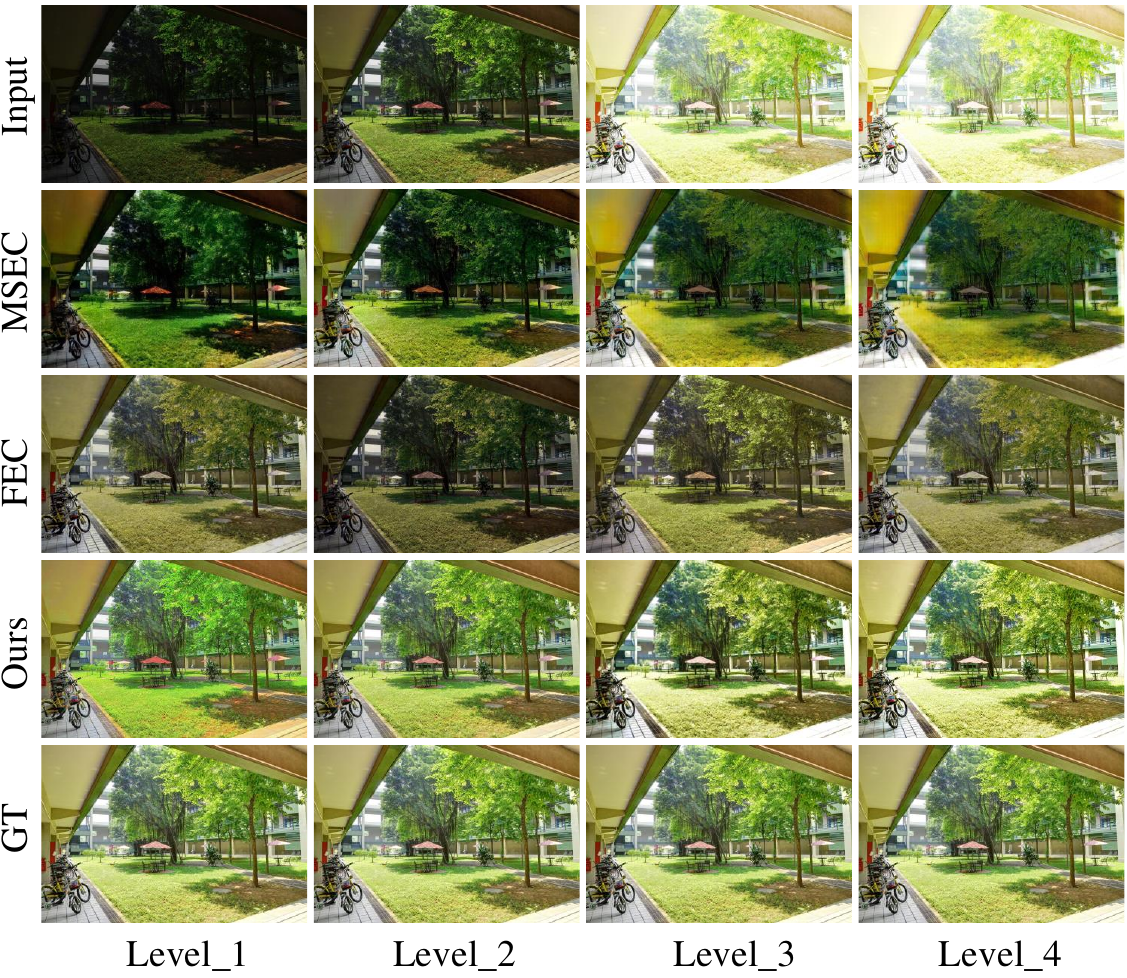}
    \caption{Visual comparison in under/overexposed scenes.}
    \label{fig:multi-exp}
    \vspace{-5.0mm}
\end{figure}

\subsection{Extending to multi-exposure correction}
In order to demonstrate the generalization performance and effectiveness of our proposed method, we compare our method with existing multiple-exposure methods (i.e., FEC \cite{huang2022deep} and MSEC \cite{afifi2021learning}) in under and over-exposure scenes. Specifically, we calculate the average brightness of each image on the V channel in HSV color space and then divide these images according to the images’ average brightness range from [0, 0.25), [0.25, 0.5), [0.5, 0.75) to [0.75, 1), denoted as Level 1, Level 2, Level 3, and Level 4, respectively. Then, we use this dataset to train FEC, MSEC, and our methods, and the results are shown in Fig. \ref{fig:multi-exp}. It is easy to observe that for underexposure Level 1 and Level 2 images, FEC and MSEC have the problem of under-enhancement, leading to the under-exposure of the enhanced image. For overexposure Level 3 and Level 4 images, FEC and MSEC have the problem of color distortion, while our method can adaptively enhance both underexposure and overexposure images.

\section{Conclusion}

In this paper, we propose a novel unsupervised recursive enhancement framework to adaptively enhance image detail and contrast in low-light images with wide dynamic ranges. Specifically, a novel ACT-Net is proposed to enhance the details and brightness of low-light images simultaneously, which utilizes the brightness adjustment branch and gradient adjustment branch to balance image contrast and detail enhancement. To adaptively enhance low-light images, a simple but effective BP-Net is proposed to perceive the brightness distribution of low-light images and predict recursive factors to control the recursive enhancement times of ACT-Net, avoiding the under or over-exposure problems. Experiments demonstrate that the proposed method achieves SOTA performance in five image quality metrics. 

In future work, we will consider adding a denoise module to remove the inevitable remaining noise in our enhanced result to further improve image quality. Besides, we will also utilize more real low-light images for training our unsupervised method to improve robustness in various low-light environments.

\bibliographystyle{IEEEtran}
\small\bibliography{reference}

\begin{thebibliography}{10}
\providecommand{\url}[1]{#1}
\csname url@samestyle\endcsname
\providecommand{\newblock}{\relax}
\providecommand{\bibinfo}[2]{#2}
\providecommand{\BIBentrySTDinterwordspacing}{\spaceskip=0pt\relax}
\providecommand{\BIBentryALTinterwordstretchfactor}{4}
\providecommand{\BIBentryALTinterwordspacing}{\spaceskip=\fontdimen2\font plus
\BIBentryALTinterwordstretchfactor\fontdimen3\font minus
  \fontdimen4\font\relax}
\providecommand{\BIBforeignlanguage}[2]{{%
\expandafter\ifx\csname l@#1\endcsname\relax
\typeout{** WARNING: IEEEtran.bst: No hyphenation pattern has been}%
\typeout{** loaded for the language `#1'. Using the pattern for}%
\typeout{** the default language instead.}%
\else
\language=\csname l@#1\endcsname
\fi
#2}}
\providecommand{\BIBdecl}{\relax}
\BIBdecl

\bibitem{huang2022deep}
J.~Huang, Y.~Liu, F.~Zhao, K.~Yan, J.~Zhang, Y.~Huang, M.~Zhou, and Z.~Xiong,
  ``Deep fourier-based exposure correction network with spatial-frequency
  interaction,'' in \emph{Computer Vision--ECCV 2022: 17th European Conference,
  Tel Aviv, Israel, October 23--27, 2022, Proceedings, Part XIX}.\hskip 1em
  plus 0.5em minus 0.4em\relax Springer, 2022, pp. 163--180.

\bibitem{ma2022toward}
L.~Ma, T.~Ma, R.~Liu, X.~Fan, and Z.~Luo, ``Toward fast, flexible, and robust
  low-light image enhancement,'' in \emph{Proceedings of the IEEE/CVF
  Conference on Computer Vision and Pattern Recognition}, 2022, pp. 5637--5646.

\bibitem{wang2020experiment}
W.~Wang, X.~Wu, X.~Yuan, and Z.~Gao, ``An experiment-based review of low-light
  image enhancement methods,'' \emph{Ieee Access}, vol.~8, pp.
  87\,884--87\,917, 2020.

\bibitem{lu2022progressive}
Y.~Lu and S.-W. Jung, ``Progressive joint low-light enhancement and noise
  removal for raw images,'' \emph{IEEE Transactions on Image Processing},
  vol.~31, pp. 2390--2404, 2022.

\bibitem{add_reference_1}
Z.~Rahman, Y.-F. Pu, M.~Aamir, and S.~Wali, ``Structure revealing of low-light
  images using wavelet transform based on fractional-order denoising and
  multiscale decomposition,'' \emph{The Visual Computer}, vol.~37, no.~5, pp.
  865--880, 2021.

\bibitem{add_reference_2}
Z.~Rahman, Z.~Ali, I.~Khan, M.~I. Uddin, Y.~Guan, and Z.~Hu, ``Diverse image
  enhancer for complex underexposed image,'' \emph{Journal of Electronic
  Imaging}, vol.~31, no.~4, pp. 041\,213--041\,213, 2022.

\bibitem{add_reference_3}
Z.~Rahman, P.~Yi-Fei, M.~Aamir, S.~Wali, and Y.~Guan, ``Efficient image
  enhancement model for correcting uneven illumination images,'' \emph{IEEE
  Access}, vol.~8, pp. 109\,038--109\,053, 2020.

\bibitem{add_reference_4}
Z.~Rahman, M.~Aamir, Z.~Ali, A.~K.~J. Saudagar, A.~AlTameem, and K.~Muhammad,
  ``Efficient contrast adjustment and fusion method for underexposed images in
  industrial cyber-physical systems,'' \emph{IEEE Systems Journal}, 2023.

\bibitem{li2021learning}
C.~Li, C.~Guo, and C.~C. Loy, ``Learning to enhance low-light image via
  zero-reference deep curve estimation,'' \emph{IEEE Transactions on Pattern
  Analysis and Machine Intelligence}, vol.~44, no.~8, pp. 4225--4238, 2021.

\bibitem{van2008visualizing}
L.~Van~der Maaten and G.~Hinton, ``Visualizing data using t-sne.''
  \emph{Journal of machine learning research}, vol.~9, no.~11, 2008.

\bibitem{wu2022uretinex}
W.~Wu, J.~Weng, P.~Zhang, X.~Wang, W.~Yang, and J.~Jiang, ``Uretinex-net:
  Retinex-based deep unfolding network for low-light image enhancement,'' in
  \emph{Proceedings of the IEEE/CVF Conference on Computer Vision and Pattern
  Recognition}, 2022, pp. 5901--5910.

\bibitem{huang2022exposure}
J.~Huang, Y.~Liu, X.~Fu, M.~Zhou, Y.~Wang, F.~Zhao, and Z.~Xiong, ``Exposure
  normalization and compensation for multiple-exposure correction,'' in
  \emph{Proceedings of the IEEE/CVF Conference on Computer Vision and Pattern
  Recognition}, 2022, pp. 6043--6052.

\bibitem{nsampi2021learning}
N.~E. Nsampi, Z.~Hu, and Q.~Wang, ``Learning exposure correction via
  consistency modeling,'' in \emph{Proc. Brit. Mach. Vision Conf.}\hskip 1em
  plus 0.5em minus 0.4em\relax BMVC, 2021.

\bibitem{yu2020searching}
Z.~Yu, C.~Zhao, Z.~Wang, Y.~Qin, Z.~Su, X.~Li, F.~Zhou, and G.~Zhao,
  ``Searching central difference convolutional networks for face
  anti-spoofing,'' in \emph{Proceedings of the IEEE/CVF Conference on Computer
  Vision and Pattern Recognition}, 2020, pp. 5295--5305.

\bibitem{zheng2022low}
S.~Zheng, Y.~Ma, J.~Pan, C.~Lu, and G.~Gupta, ``Low-light image and video
  enhancement: A comprehensive survey and beyond,'' \emph{arXiv preprint
  arXiv:2212.10772}, 2022.

\bibitem{liu2021benchmarking}
J.~Liu, D.~Xu, W.~Yang, M.~Fan, and H.~Huang, ``Benchmarking low-light image
  enhancement and beyond,'' \emph{International Journal of Computer Vision},
  vol. 129, pp. 1153--1184, 2021.

\bibitem{li2021low}
C.~Li, C.~Guo, L.~Han, J.~Jiang, M.-M. Cheng, J.~Gu, and C.~C. Loy, ``Low-light
  image and video enhancement using deep learning: A survey,'' \emph{IEEE
  transactions on pattern analysis and machine intelligence}, vol.~44, no.~12,
  pp. 9396--9416, 2021.

\bibitem{jiang2022degrade}
K.~Jiang, Z.~Wang, Z.~Wang, C.~Chen, P.~Yi, T.~Lu, and C.-W. Lin, ``Degrade is
  upgrade: Learning degradation for low-light image enhancement,'' in
  \emph{Proceedings of the AAAI Conference on Artificial Intelligence},
  vol.~36, no.~1, 2022, pp. 1078--1086.

\bibitem{jiang2022unsupervised}
Q.~Jiang, Y.~Mao, R.~Cong, W.~Ren, C.~Huang, and F.~Shao, ``Unsupervised
  decomposition and correction network for low-light image enhancement,''
  \emph{IEEE Transactions on Intelligent Transportation Systems}, vol.~23,
  no.~10, pp. 19\,440--19\,455, 2022.

\bibitem{pizer1987adaptive}
S.~M. Pizer, E.~P. Amburn, J.~D. Austin, R.~Cromartie, A.~Geselowitz, T.~Greer,
  B.~ter Haar~Romeny, J.~B. Zimmerman, and K.~Zuiderveld, ``Adaptive histogram
  equalization and its variations,'' \emph{Computer vision, graphics, and image
  processing}, vol.~39, no.~3, pp. 355--368, 1987.

\bibitem{abdullah2007dynamic}
M.~Abdullah-Al-Wadud, M.~H. Kabir, M.~A.~A. Dewan, and O.~Chae, ``A dynamic
  histogram equalization for image contrast enhancement,'' \emph{IEEE
  Transactions on Consumer Electronics}, vol.~53, no.~2, pp. 593--600, 2007.

\bibitem{reza2004realization}
A.~M. Reza, ``Realization of the contrast limited adaptive histogram
  equalization (clahe) for real-time image enhancement,'' \emph{Journal of VLSI
  signal processing systems for signal, image and video technology}, vol.~38,
  no.~1, pp. 35--44, 2004.

\bibitem{lee2007efficient}
S.~Lee, ``An efficient content-based image enhancement in the compressed domain
  using retinex theory,'' \emph{IEEE transactions on circuits and systems for
  video technology}, vol.~17, no.~2, pp. 199--213, 2007.

\bibitem{land1977retinex}
E.~H. Land, ``The retinex theory of color vision,'' \emph{Scientific american},
  vol. 237, no.~6, pp. 108--129, 1977.

\bibitem{cai2017joint}
B.~Cai, X.~Xu, K.~Guo, K.~Jia, B.~Hu, and D.~Tao, ``A joint intrinsic-extrinsic
  prior model for retinex,'' in \emph{Proceedings of the IEEE international
  conference on computer vision}, 2017, pp. 4000--4009.

\bibitem{add_reference_5}
Z.~Rahman, J.~A. Bhutto, M.~Aamir, Z.~A. Dayo, and Y.~Guan, ``Exploring a
  radically new exponential retinex model for multi-task environments,''
  \emph{Journal of King Saud University-Computer and Information Sciences},
  vol.~35, no.~7, p. 101635, 2023.

\bibitem{wei2018deep}
C.~Wei, W.~Wang, W.~Yang, and J.~Liu, ``Deep retinex decomposition for
  low-light enhancement,'' \emph{arXiv preprint arXiv:1808.04560}, 2018.

\bibitem{wang2022local}
H.~Wang, K.~Xu, and R.~W. Lau, ``Local color distributions prior for image
  enhancement,'' in \emph{Computer Vision--ECCV 2022: 17th European Conference,
  Tel Aviv, Israel, October 23--27, 2022, Proceedings, Part XVIII}.\hskip 1em
  plus 0.5em minus 0.4em\relax Springer, 2022, pp. 343--359.

\bibitem{dudhane2022burst}
A.~Dudhane, S.~W. Zamir, S.~Khan, F.~S. Khan, and M.-H. Yang, ``Burst image
  restoration and enhancement,'' in \emph{Proceedings of the IEEE/CVF
  Conference on Computer Vision and Pattern Recognition}, 2022, pp. 5759--5768.

\bibitem{dong2022abandoning}
X.~Dong, W.~Xu, Z.~Miao, L.~Ma, C.~Zhang, J.~Yang, Z.~Jin, A.~B.~J. Teoh, and
  J.~Shen, ``Abandoning the bayer-filter to see in the dark,'' in
  \emph{Proceedings of the IEEE/CVF Conference on Computer Vision and Pattern
  Recognition}, 2022, pp. 17\,431--17\,440.

\bibitem{zhang2022deep}
Z.~Zhang, H.~Zheng, R.~Hong, M.~Xu, S.~Yan, and M.~Wang, ``Deep color
  consistent network for low-light image enhancement,'' in \emph{Proceedings of
  the IEEE/CVF Conference on Computer Vision and Pattern Recognition}, 2022,
  pp. 1899--1908.

\bibitem{zheng2022semantic}
S.~Zheng and G.~Gupta, ``Semantic-guided zero-shot learning for low-light
  image/video enhancement,'' in \emph{Proceedings of the IEEE/CVF Winter
  Conference on Applications of Computer Vision}, 2022, pp. 581--590.

\bibitem{wang2019underexposed}
R.~Wang, Q.~Zhang, C.-W. Fu, X.~Shen, W.-S. Zheng, and J.~Jia, ``Underexposed
  photo enhancement using deep illumination estimation,'' in \emph{Proceedings
  of the IEEE/CVF conference on computer vision and pattern recognition}, 2019,
  pp. 6849--6857.

\bibitem{jiang2021enlightengan}
Y.~Jiang, X.~Gong, D.~Liu, Y.~Cheng, C.~Fang, X.~Shen, J.~Yang, P.~Zhou, and
  Z.~Wang, ``Enlightengan: Deep light enhancement without paired supervision,''
  \emph{IEEE transactions on image processing}, vol.~30, pp. 2340--2349, 2021.

\bibitem{afifi2021learning}
M.~Afifi, K.~G. Derpanis, B.~Ommer, and M.~S. Brown, ``Learning multi-scale
  photo exposure correction,'' in \emph{Proceedings of the IEEE/CVF Conference
  on Computer Vision and Pattern Recognition}, 2021, pp. 9157--9167.

\bibitem{nsamp2018learning}
N.~Nsamp, Z.~Hu, and Q.~Wang, ``Learning exposure correction via consistency
  modeling,'' in \emph{Proceedings of the British Machine Vision Conference
  (BMVC)}, 2018, pp. 1--12.

\bibitem{cai2018learning}
J.~Cai, S.~Gu, and L.~Zhang, ``Learning a deep single image contrast enhancer
  from multi-exposure images,'' \emph{IEEE Transactions on Image Processing},
  vol.~27, no.~4, pp. 2049--2062, 2018.

\bibitem{bychkovsky2011learning}
V.~Bychkovsky, S.~Paris, E.~Chan, and F.~Durand, ``Learning photographic global
  tonal adjustment with a database of input/output image pairs,'' in \emph{CVPR
  2011}.\hskip 1em plus 0.5em minus 0.4em\relax IEEE, 2011, pp. 97--104.

\bibitem{wang2013naturalness}
S.~Wang, J.~Zheng, H.-M. Hu, and B.~Li, ``Naturalness preserved enhancement
  algorithm for non-uniform illumination images,'' \emph{IEEE transactions on
  image processing}, vol.~22, no.~9, pp. 3538--3548, 2013.

\bibitem{LIME}
X.~Guo, Y.~Li, and H.~Ling, ``Lime: Low-light image enhancement via
  illumination map estimation,'' \emph{IEEE Transactions on image processing},
  vol.~26, no.~2, pp. 982--993, 2016.

\bibitem{MEF}
K.~Ma, K.~Zeng, and Z.~Wang, ``Perceptual quality assessment for multi-exposure
  image fusion,'' \emph{IEEE Transactions on Image Processing}, vol.~24,
  no.~11, pp. 3345--3356, 2015.

\bibitem{DICM}
C.~Lee, C.~Lee, and C.-S. Kim, ``Contrast enhancement based on layered
  difference representation,'' in \emph{2012 19th IEEE international conference
  on image processing}.\hskip 1em plus 0.5em minus 0.4em\relax IEEE, 2012, pp.
  965--968.

\bibitem{zhang2020self}
Y.~Zhang, X.~Di, B.~Zhang, and C.~Wang, ``Self-supervised image enhancement
  network: Training with low light images only,'' \emph{arXiv preprint
  arXiv:2002.11300}, 2020.

\bibitem{liu2021retinex}
R.~Liu, L.~Ma, J.~Zhang, X.~Fan, and Z.~Luo, ``Retinex-inspired unrolling with
  cooperative prior architecture search for low-light image enhancement,'' in
  \emph{Proceedings of the IEEE/CVF Conference on Computer Vision and Pattern
  Recognition}, 2021, pp. 10\,561--10\,570.

\bibitem{zhang2021beyond}
Y.~Zhang, X.~Guo, J.~Ma, W.~Liu, and J.~Zhang, ``Beyond brightening low-light
  images,'' \emph{International Journal of Computer Vision}, vol. 129, pp.
  1013--1037, 2021.

\bibitem{yang2020fidelity}
W.~Yang, S.~Wang, Y.~Fang, Y.~Wang, and J.~Liu, ``From fidelity to perceptual
  quality: A semi-supervised approach for low-light image enhancement,'' in
  \emph{Proceedings of the IEEE/CVF conference on computer vision and pattern
  recognition}, 2020, pp. 3063--3072.

\bibitem{wang2022low}
Y.~Wang, R.~Wan, W.~Yang, H.~Li, L.-P. Chau, and A.~Kot, ``Low-light image
  enhancement with normalizing flow,'' in \emph{Proceedings of the AAAI
  Conference on Artificial Intelligence}, vol.~36, no.~3, 2022, pp. 2604--2612.

\bibitem{huynh2008scope}
Q.~Huynh-Thu and M.~Ghanbari, ``Scope of validity of psnr in image/video
  quality assessment,'' \emph{Electronics letters}, vol.~44, no.~13, pp.
  800--801, 2008.

\bibitem{wang2004image}
Z.~Wang, A.~C. Bovik, H.~R. Sheikh, and E.~P. Simoncelli, ``Image quality
  assessment: from error visibility to structural similarity,'' \emph{IEEE
  transactions on image processing}, vol.~13, no.~4, pp. 600--612, 2004.

\bibitem{zhang2018unreasonable}
R.~Zhang, P.~Isola, A.~A. Efros, E.~Shechtman, and O.~Wang, ``The unreasonable
  effectiveness of deep features as a perceptual metric,'' in \emph{Proceedings
  of the IEEE conference on computer vision and pattern recognition}, 2018, pp.
  586--595.

\bibitem{mittal2012making}
A.~Mittal, R.~Soundararajan, and A.~C. Bovik, ``Making a “completely blind”
  image quality analyzer,'' \emph{IEEE Signal processing letters}, vol.~20,
  no.~3, pp. 209--212, 2012.

\bibitem{agaian2000new}
S.~S. Agaian, K.~Panetta, and A.~M. Grigoryan, ``A new measure of image
  enhancement,'' in \emph{IASTED International Conference on Signal Processing
  \& Communication}.\hskip 1em plus 0.5em minus 0.4em\relax Citeseer, 2000, pp.
  19--22.

\end{thebibliography}

\end{document}